\documentclass[lettersize,journal]{IEEEtran}
\usepackage{times}
\usepackage{graphicx}
\usepackage{amsmath}

\usepackage[
    backend=bibtex,
    style=authoryear,
    maxcitenames=2,
    mincitenames=1,
    minbibnames=1,
    maxbibnames=6,
    url=false,
    isbn=false,
    doi=false,
    eprint=false,
    giveninits=true
]{biblatex}  
\usepackage{multicol}
\usepackage[bookmarks=true]{hyperref}
\usepackage[caption=false,font=footnotesize,labelfont=sf,textfont=sf]{subfig}

\usepackage{svg}

\usepackage{algorithm}
\usepackage{algpseudocode}

\usepackage{xcolor, soul}
\sethlcolor{yellow}

\usepackage{xspace}
\usepackage{multirow}

\usepackage{enumitem}
\usepackage{adjustbox}
\usepackage{threeparttable}
\usepackage{array, multirow}        

\newcolumntype{C}[1]{>{\centering\let\newline\\\arraybackslash\hspace{0pt}}m{#1}}

\usepackage[normalem]{ulem}

\newcommand{\InitialTraj}{\ensuremath{\xi_{0}}\xspace}
\newcommand{\Traj}[1]{\ensuremath{\xi_{#1}}\xspace}
\newcommand{\OptimalTraj}{\ensuremath{\xi^{*}}\xspace}
\newcommand{\LangCorrUttered}[2]{\ensuremath{l_{#1}^{#2}}\xspace}
\newcommand{\FeatureSingle}[1]{\ensuremath{\phi_{#1}}\xspace}
\newcommand{\FeatureSet}{\ensuremath{\Phi}\xspace}
\newcommand{\OptimalFeature}{\ensuremath{\phi^{*}}\xspace}

\newcommand{\Env}[2]{\ensuremath{E_{#1}^{#2}}\xspace}
\newcommand{\DeformationFn}{\ensuremath{\delta}\xspace}
\newcommand{\Force}{\ensuremath{F}\xspace}
\newcommand{\ObjPos}[1]{\ensuremath{o_{pos}^{#1}}\xspace}
\newcommand{\ObjName}[1]{\ensuremath{o_{name}^{#1}}\xspace}
\newcommand{\conceptNameS}{feature\xspace}
\newcommand{\conceptNameP}{features\xspace}
\newcommand{\LanguagePhraseSet}{\ensuremath{T_{\phi}}\xspace}
\newcommand{\LanguagePhraseSingle}{\ensuremath{t_{\phi}}\xspace}

\newcommand{\Embedding}[1]{\ensuremath{q(#1)}\xspace}

\usepackage{xcolor}

\begin{document}

\title{ExTraCT -- Explainable Trajectory Corrections from language inputs using Textual description of features}
\author{
    J-Anne Yow\,$^{1,2*}$, Neha Priyadarshini Garg\,$^{1}$, Manoj Ramanathan\,$^{1}$ and Wei Tech Ang\,$^{1,2}$
    \thanks{$^{1}$Rehabalitation Research Institute of Singapore (RRIS), Clinical Science Building, 308232 Singapore. RRIS is a joint research institute by Nanyang Technological University (NTU), Agency for Science, Technology and Research (A*STAR) and National Healthcare Group (NHG), Singapore}
    
    \thanks{$^{2}$Singapore-ETH Centre, Future Health Technologies Programme, CREATE Tower, Singapore}

    \thanks{Correspondence$^{*}$: J-Anne Yow (janne.yow@ntu.edu.sg)}
    
    \thanks{
    This work is supported by the Rehabilitation Research Institute of Singapore and the National Research Foundation, Prime Minister's Office, Singapore, under its Campus for Research Excellence and Technological Enterprise (CREATE) programme.
    }

}

\maketitle

\begin{abstract}

Natural language provides an intuitive and expressive way of conveying human intent to robots. Prior works employed end-to-end methods for learning trajectory deformations from language corrections. However, such methods do not generalize to new initial trajectories or object configurations. 
This work presents ExTraCT, a modular framework for trajectory corrections using natural language that combines Large Language Models (LLMs) for natural language understanding and trajectory deformation functions.
Given a scene, ExTraCT generates the trajectory modification features (scene-specific and scene-independent) and their corresponding natural language textual descriptions for the objects in the scene online based on a template. We use LLMs for semantic matching of user utterances to the textual descriptions of features. Based on the feature matched, a trajectory modification function is applied to the initial trajectory, allowing generalization to unseen trajectories and object configurations. 
Through user studies conducted both in simulation and with a physical robot arm, we demonstrate that trajectories deformed using our method were more accurate and were preferred in about 80\% of cases, outperforming the baseline. We also showcase the versatility of our system in a manipulation task and an assistive feeding task.

\end{abstract}



\IEEEpeerreviewmaketitle

\section{Introduction}
\label{sec:intro}

Motion planning algorithms optimize trajectories based on predefined cost functions, which usually consider robot dynamics and environmental constraints. However, robots need to account for human preferences to assist effectively when working with humans. Human preferences can vary based on the environment and human factors; e.g. while throwing trash, humans might want the robot to avoid food items, or when placing a cup, they may prefer the robot to stay close to the table and avoid other objects. Thus, incorporating all possible human preferences in the cost function is challenging. To address this, we explore the problem of natural language trajectory corrections, i.e. modifying a robot's initial trajectory based on natural language corrections provided by a human. We choose natural language as it provides an intuitive and expressive way for humans to convey their preferences.

\begin{figure*}[!t]
\centering
\includegraphics[width=0.85\textwidth
]{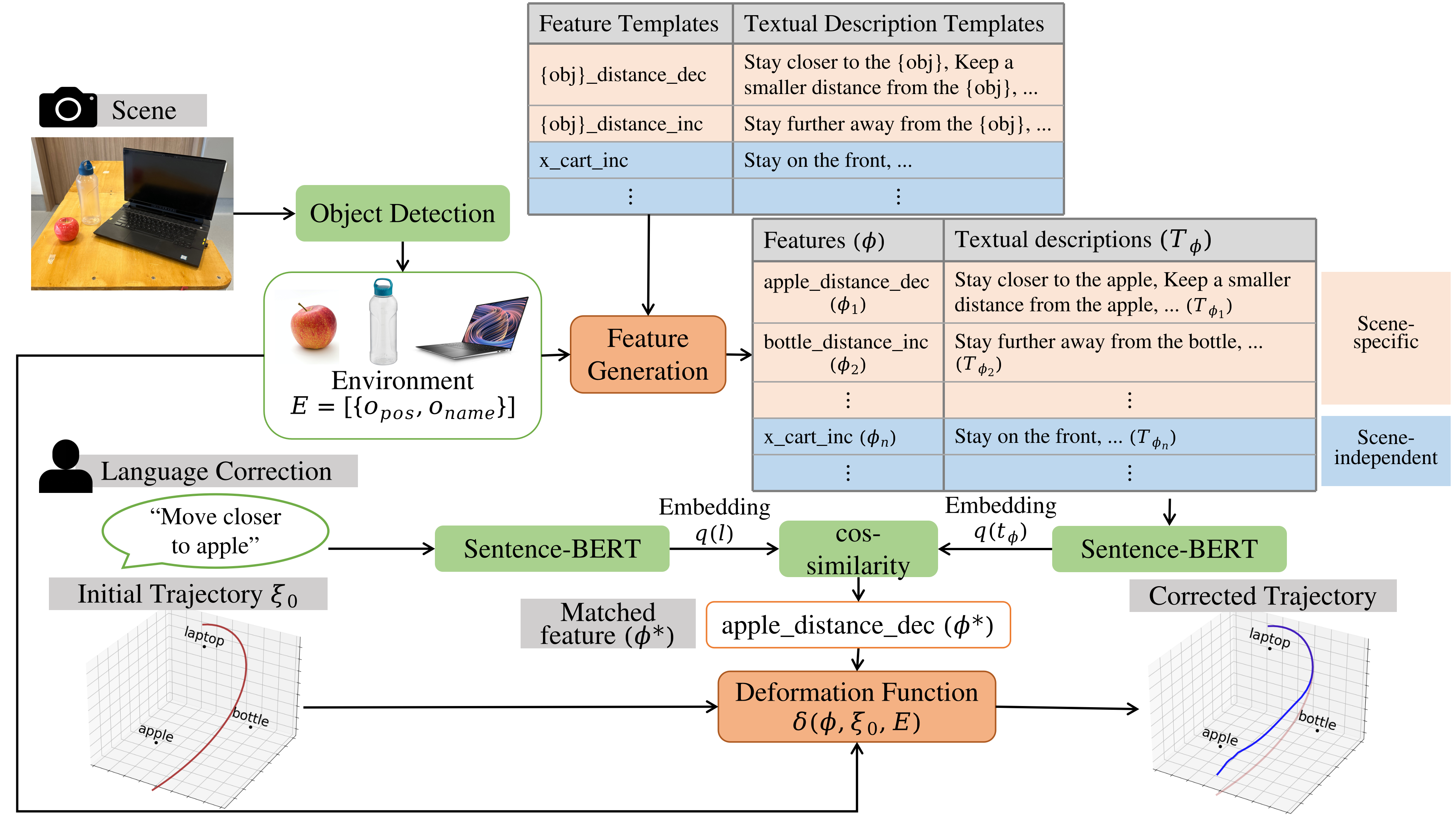}
\vspace{-5pt}
\caption{Architecture of ExTraCT. Given the objects in a scene, the features \FeatureSingle{} and corresponding textual descriptions \LanguagePhraseSet are generated online. We obtain the embeddings of the language correction \Embedding{\LangCorrUttered{}{}} and the phrases ($\LanguagePhraseSingle \in \LanguagePhraseSet$) in the textual descriptions of the features \Embedding{\LanguagePhraseSingle}, and use semantic textual similarity to obtain the most similar textual description, which is mapped to feature \OptimalFeature. A deformation function \DeformationFn is used to deform the initial trajectory \InitialTraj based on the feature \OptimalFeature and the object positions in the environment \Env{}{}. A trajectory optimizer is used to ensure that the robot's kinematic constraints are satisfied.}
\label{fig:architecture}
\end{figure*}



A key challenge in natural language trajectory corrections is 
mapping the natural language to the robot action, which is the deformed trajectory.
Existing works \parencite{broad2017real, sharma2022correcting, bucker2022reshaping, bucker2022latte} try to learn a direct mapping between natural language and robot trajectories or actions using offline training paradigms. To improve generalization to different objects and phrases, some works \parencite{sharma2022correcting, bucker2022reshaping, bucker2022latte} leverage on foundational models, i.e. BERT \parencite{devlin2018bert} and CLIP \parencite{radford2021learning}. However, these models struggle to generalize to varied initial trajectories and object poses due to dataset limitations.
Furthermore, with a direct mapping learnt between natural language and robot action, it is difficult to understand the root cause of failures, which could stem from issues in language grounding, scene understanding, or inaccuracies in trajectory deformation function.

To address these limitations, we separate language understanding from trajectory deformation, thus enabling a more accurate interpretation of instructions. First, we match the language corrections to a short description of the change in trajectory, which we term a feature. We define a set of feature templates and their corresponding textual description templates, which are then used to generate features and their textual descriptions (Fig. \ref{fig:architecture}) for any given scene online. 
The language uttered by the user can then be mapped to the most likely feature by computing the semantic similarity between the textual descriptions of the feature and the correction uttered by the user using Large Language Models \parencite{devlin2018bert, wang2020minilm, brown2020gpt3, chowdhery2022palm}. 
For features that are mapped with insufficient confidence (\(\le 0.6\)), our approach can alert the end-user instead of generating a random modified trajectory.

Once the most likely feature is determined, the initial trajectory can be modified based on a trajectory deformation function, allowing generalization to different object configurations and trajectories. 

This separation of language understanding from trajectory deformation allows for more precise and context-aware robotic responses. By decoupling these two elements, our method can easily expand to new tasks, as shown in Section \ref{sec:feeding}. Furthermore, it provides clearer insights into potential sources of failure, as it distinguishes between errors in language interpretation and trajectory execution. Thus, our approach improves the accuracy and generalization of language corrections and enhances the interpretability and reliability of robotic systems in executing these corrections.

While our approach is limited by the feature types in the feature templates, end-to-end training methods would encounter similar limitations when dealing with unseen features. However, they are additionally limited by the object configurations and initial trajectories used in the training data, as shown in Section \ref{sec:baseline}. Our formulation can generalize to different object configurations and initial trajectories without training a model. 
Our key insight is that integrating the strengths of LLMs in handling language diversity with a hand-crafted approach for trajectory modifications bypasses the need for end-to-end training while achieving comparable or better performance. Furthermore, end-to-end training methods are data-intensive. Collecting data is challenging in the robotics domain, while generating data would require a similar amount of hand-crafted features as our proposed approach.


We evaluated our system through within-subject user studies in simulation and the real world. 
Our results show that our method had higher accuracy and was rated higher in approximately $80\%$ of test cases compared to the state-of-the-art method LaTTe \parencite{bucker2022latte}, which also uses LLMs but was trained in an end-to-end fashion.
We further analysed the failure cases of our method and showed that our method could be improved further by adding more phrases to the textual description templates. 

Our contributions in this work are threefold: (1) We introduce a modular framework that integrates LLMs with trajectory deformation functions for trajectory corrections using language without end-to-end training. (2) We conducted extensive quantitative experiments on a substantial dataset, complemented by user studies comparing our method against a baseline that deforms trajectories based on language in an end-to-end fashion. (3) We demonstrate the versatility of our framework through its application to a range of tasks, including general object manipulation and assistive feeding.

\section{Related Work}
\label{sec:related_work}
There are various ways of conveying human preference to the robot, such as through language \parencite{bucker2022reshaping, bucker2022latte, sharma2022correcting}, physical interaction \parencite{bajcsy2017learning, bajcsy2018learning, bobu2021feature}, rankings \parencite{jain2015learning} and joystick inputs \parencite{spencer2020learning}. In this work, we use language as it is the most natural and intuitive way of communicating human preference \parencite{tellex2020robots}.

Language correction works can be broadly classified into two categories -- generating new trajectories \parencite{broad2017real, sharma2022correcting, liang2023code, zha2023distilling} and modifying existing ones \parencite{bucker2022reshaping, bucker2022latte}. 
The first category, trajectory generation, involves creating new motion plans to enable robots to correct errors to complete tasks.
Consider a task where a robot provides feeding assistance to a user. An online correction such as ``move 5cm to the left'' when acquiring the food might be provided so the robot can align more accurately with the food morsel before acquiring it. This directs the robot to generate a new trajectory to move to the left for more precise alignment.
On the other hand, our work focuses on the second category of corrections, which involves modifying an existing trajectory. This includes relative corrections such as ``scoop a larger amount of food''. Such corrections require adjustments to an existing trajectory, requiring an understanding of the initial plan or trajectory. Modifying existing trajectories is important, as human preferences are commonly expressed in relative terms. 



Both categories of language corrections face a common challenge: translating natural language to robot actions, also known as language grounding. Langauge grounding can be categorized into three approaches -- semantic parsing to probabilistic graphs, end-to-end learning using embeddings and prompting LLMs to generate code.

\subsubsection{Semantic Parsing to Probabilistic Graphs}
Earlier works in language correction use a grammatical structure to represent language and ground language by learning the weights of functions of factors \parencite{broad2017real}. A probabilistic graphical model (distributed correspondence graph, DCG) is used to ground language to a set of features that relate to the environment and context. Each feature represents a cost or constraint, which a motion planner then optimizes. Given the variety of possible phrases a human can provide to correct the same feature, grounding language using a fixed grammatical structure limits the generalization of this approach.
We differ from this approach in how we ground language. Instead of parsing language corrections to phrases with a fixed grammar, we leverage the textual embeddings in LLMs \parencite{devlin2018bert, wang2020minilm, brown2020gpt3, chowdhery2022palm} to ground them to textual descriptions of features. 


\subsubsection{End-to-End Learning Using Embeddings}
Prior works have explored LLMs, combining BERT \parencite{devlin2018bert} (textual) embeddings and CLIP \parencite{radford2021learning} (textual and visual) embeddings to align visual and language representations. 
Sharma \textit{et. al.} \parencite{sharma2022correcting} learn a 2D cost map from CLIP and BERT embeddings, which converts language corrections to a cost map that is optimized using a motion planner to obtain the corrected trajectory.
Bucker \textit{et. al.} goes a step further by directly learning a corrected trajectory from CLIP and BERT embeddings in 2D \parencite{bucker2022reshaping} and 3D \parencite{bucker2022latte}. 
Additionally, a transformer encoder obtains geometric embeddings of the object poses and an initial trajectory. A transformer decoder combines textual, visual and geometric embeddings to generate the corrected trajectory. 
Geometric embeddings have also been combined with textual embeddings to learn a robot policy conditioned on language \parencite{cui2023no}, but a shared autonomy paradigm was used to reduce the complexity.
These approaches deploy an end-to-end learning approach requiring large multi-modal datasets, including image, text, and trajectory data, which are difficult to obtain. 
Compared to these works, our proposed approach leverages LLMs to summarize natural language corrections into concise and informative textual representations, thus removing the need for multi-modal training data to ground language corrections.


\subsubsection{Prompting LLMs to Generate Code}
Concurrent with our work, advances in Large Language Models (LLMs) have led to recent works \parencite{liang2023code, huang2023voxposer, yu2023language, zha2023distilling} employing LLMs to generate executable code from natural language instructions. These works mainly focus on generating robot plans or trajectories given a phrase, either by calling pre-defined motion primitives \parencite{liang2023code, zha2023distilling} or by designing specific reward functions \parencite{huang2023voxposer, yu2023language}. This approach has shown versatility in handling diverse instructions and constructing sequential policy logic. 
However, the adoption of these methods is hindered by significant computational costs due to the need for larger, general-purpose models like GPT-4, which cannot be run on standard commercial hardware. Remote execution via API calls is a viable option, but this is limited by the need for a stable internet connection and the rate limits imposed on the APIs, which can hinder real-time applications.
Moreover, similar to our approach that integrates hand-crafted templates, the effective use of LLMs for specific task-oriented code generation also requires many in-context examples.  
In contrast, our approach provides a more cost-effective solution for language understanding in robotic systems by using semantic textual similarity, eliminating the need for extensive prompting or high computational demands. 
Although our method requires additional modules to process complex language utterances, such as referring expressions and compound sentences, it is a more viable alternative in settings constrained by limited computational resources and internet connectivity. This makes our framework well-suited for practical applications where efficiency is a key consideration.

\section{Approach}
\label{sec:approach}

\subsection{Problem Definition}
\label{sec:problem}

Our goal is to develop an interface that allows users to modify the trajectory of robot manipulators based on their preferences conveyed through natural language corrections. 
More specifically, the problem can be described as finding the most likely trajectory \OptimalTraj given the environment \Env{}{}, the language correction \LangCorrUttered{}{} provided by the human and an initial trajectory \InitialTraj.

\begin{equation}
    \label{eqn_prob_traj}
    \OptimalTraj = \underset{\Traj{}}{\mathrm{argmax}}\, P(\Traj{} | \Env{}{}, \LangCorrUttered{}{}, \InitialTraj)
\end{equation}

The environment consists of a set of objects, with each object having two attributes -- object name \ObjName{} and object position \ObjPos{}, i.e. $\Env{}{} = \{(\ObjName{i}, \ObjPos{i})\} $. 
The object names and poses in the environment can be obtained using perception algorithms (e.g. Mask R-CNN \parencite{he2017mask}, YOLO \parencite{redmon2018yolov3}) or foundational models (e.g. OWL-ViT \parencite{minderer2022simple}, Grounding DINO \parencite{liu2023grounding}).
In our work, we used Mask R-CNN to get the environment \Env{}{}. 

\subsection{Features}
\label{sec: feature_space}
The space of possible trajectories \Traj{} is infinite. However, realistically, it can be bounded by the environment and motion planner constraints \parencite{broad2017real}. Therefore, we assume that the language correction \LangCorrUttered{}{} can be mapped to a finite set of \conceptNameP \FeatureSet that can be scene-specific or scene-independent. 
Each \conceptNameS $\FeatureSingle{} \in \FeatureSet$ corresponds to a deformed trajectory, i.e. $\Traj{} = \DeformationFn(\FeatureSingle{}, \InitialTraj, \Env{}{})$ where \DeformationFn is the trajectory deformation function which we describe in Section \ref{sec: deformations}. Thus, we can obtain the most likely trajectory $\OptimalTraj = \DeformationFn(\OptimalFeature, \InitialTraj, \Env{}{})$ from the most likely feature \OptimalFeature \footnote{We assume that a given language correction will correspond to only one feature. Practically, a complex language correction can map to multiple features. We will address this issue in the future.}. This reduces the problem of finding the most likely trajectory \OptimalTraj to the problem of finding the most likely feature \OptimalFeature. Thus, our problem can be rewritten as:

\begin{equation} 
    \label{eqn_prob_ft}
    \begin{split}
        \OptimalFeature
        & = \underset{\FeatureSingle{} \in \FeatureSet}{\mathrm{argmax}}\, P(\FeatureSingle{} |  \LangCorrUttered{}{}) \\
    \end{split}
\end{equation}

 
where $P(\FeatureSingle{} |  \LangCorrUttered{}{})$ is the probability of feature \FeatureSingle{} given language utterance \LangCorrUttered{}{}. The features can be categorized into two types -- scene-specific and scene-independent features. Scene-specific features depend on the objects in the scene, while scene-independent features are not.

As proof of concept, we define 2 scene-specific and 6 scene-independent feature templates (Table \ref{table:features}). Scene-specific features are generated online for each object detected in the scene, allowing the approach to generalize to different objects. They are defined based on \textit{object distance}, which is to either increase (\texttt{obj\_distance\_increase}) or decrease (\texttt{obj\_distance\_decrease}) the distance to an object. 
Scene-independent features are based on the 3 axes in \textit{cartesian} space, i.e. to move the gripper up (\texttt{z\_cart\_increase}), down (\texttt{z\_cart\_decrease}), left (\texttt{y\_cart\_decrease}), right (\texttt{y\_cart\_increase}), forward (\texttt{x\_cart\_increase}) and backward (\texttt{x\_cart\_decrease}). The space of feature templates is expandable and can be made specific to the robot's task, as demonstrated in Section \ref{sec:feeding}.


\begin{table}[!h]
\caption{Feature Templates (FT) and their Textual Description Templates (TDT)}
\begin{center}
\begin{tabular}{ | m{0.5cm} | m{3.5cm} | m{3.5cm} | } 
    \hline
    FT & {obj\_distance\_decrease} & {obj\_distance\_increase}\\ 
    \hline
    \multirow{5}{*}{TDT} & Move closer to \{obj\} & Move further away from \{obj\} \\ 
        & \textbf{Stay close to \{obj\}} & \textbf{Stay away from \{obj\}} \\ 
        & \textbf{Decrease distance to \{obj\}} & \textbf{Increase distance to \{obj\}} \\
        & Keep a smaller distance from \{obj\} & Keep a bigger distance from \{obj\} \\
        & & Avoid \{obj\} \\
    \hline
    
    \multicolumn{3}{c}{} \\[-0.5em] 
    
    \hline
    FT & {z\_cart\_decrease} & {z\_cart\_increase} \\
    \hline
    \multirow{9}{*}{TDT} & Move closer to table & Move further away from table \\
    & \textbf{Stay closer to table} & \textbf{Stay away from table} \\
    & Move lower & Move higher \\
    & Move down & Move up \\
    & \textbf{Stay down} & \textbf{Stay up} \\
    & Go to the bottom & Stay on the upper part \\
    & Down & Up \\
    & Low & Top \\
    & & Go to the top \\
    \hline
    
    \multicolumn{3}{c}{} \\[-0.5em]
    
    \hline
    FT & y\_cart\_decrease & y\_cart\_increase \\
    \hline
    \multirow{5}{*}{TDT} & \textbf{Stay on the left} & \textbf{Stay on the right} \\
    & \textbf{Go to the left} & \textbf{Go to the right} \\
    & Move left & Move right \\
    & Move more towards the left & Move more towards the right \\
    & Left & Right \\
    \hline

    \multicolumn{3}{c}{} \\[-0.5em]
    
    \hline
    FT & x\_cart\_decrease & x\_cart\_increase \\
    \hline
    \multirow{8}{*}{TDT}& \textbf{Stay at the back} & \textbf{Stay at the front} \\
    & \textbf{Go to the back} & \textbf{Go to the front} \\
    & Move back & Move front \\
    & Stay back & Stay front \\
    & Move backward & Move forward \\
    & Go behind & \\
    \hline
\end{tabular}
\begin{tablenotes}
    Only the bolded phrases were used for the analysis in Section \ref{sec: effect-phrases}.
\end{tablenotes}
\end{center}
\label{table:features}
\end{table}

\subsection{Textual Descriptions and Optimal Feature Selection}
To compute $P(\FeatureSingle{} | \LangCorrUttered{}{})$, we generate a textual description \LanguagePhraseSet for each \conceptNameS \FeatureSingle{}. Each textual description consists of a set of language phrases, which includes commonly used phrases to modify the trajectory for a particular feature.
Table \ref{table:features} shows the textual description templates (TDT) for each feature template (FT). 
During feature generation for a scene, the \{obj\} placeholder in scene-specific feature templates will be replaced by the object name \ObjName{} obtained using object detection, as shown in Fig. \ref{fig:architecture}. 

We initially define a few phrases in the textual description templates. 
If feature matching fails due to out-of-distribution user utterances, additional phrases can be added to the textual description templates easily to improve feature matching.
We examine how the phrases in the textual description templates can affect feature matching and performance in Section \ref{sec: effect-phrases}.



Since there is a one-to-one mapping between the feature and its textual description, Eq. \ref{eqn_prob_ft} can be rewritten as:

 \begin{equation} \label{eqn_language_phrases}
\begin{split}
    \OptimalFeature
    & = \underset{\FeatureSingle{} \in \FeatureSet}{\mathrm{argmax}}\, P(\LanguagePhraseSet | \LangCorrUttered{}{}) \\
\end{split}
\end{equation}






To compute $P(\LanguagePhraseSet | \LangCorrUttered{}{})$, we leverage large language models (LLMs) to map diverse language phrases to fixed-length vectors called embeddings. To capture the semantic meaning of sentences, we used Sentence Transformers \parencite{reimers_2019_sentence_bert}, which is fine-tuned for semantic similarity tasks. We chose the pre-trained \texttt{all-MiniLM-L6-v2} model \parencite{wang2020minilm} provided by Sentence Transformers \footnote{\url{https://www.sbert.net/docs/pretrained_models.html}} as it provided us with the best trade-off between speed and performance.
Semantically closer language phrases are more likely to have higher cosine similarity between their embeddings. 
Thus, $P(\LanguagePhraseSet | \LangCorrUttered{}{})$ can be defined as follows:

\begin{equation} \label{eq_cos_similarity}
    P(\LanguagePhraseSet | \LangCorrUttered{}{}) \propto \underset{\LanguagePhraseSingle \in \LanguagePhraseSet}{\mathrm{max}}\,
     \Embedding{\LanguagePhraseSingle}.\Embedding{\LangCorrUttered{}{}}  /||\Embedding{\LanguagePhraseSingle}||. ||\Embedding{\LangCorrUttered{}{}}||
\end{equation}

where \Embedding{x} is the embedding for a language phrase x.
Once the most likely \conceptNameS \OptimalFeature is obtained, the trajectory \OptimalTraj can be obtained using a deformation function.

\subsection{Deformation Function}
\label{sec: deformations}
A deformation function $\DeformationFn(\FeatureSingle{}, \InitialTraj, \Env{}{})$ modifies a trajectory based on a feature \FeatureSingle{}. 
First, we calculate the force \Force to be exerted on each waypoint of the trajectory. 

For scene-specific features, i.e. \textit{object distance} features, the force exerted is dependent on the object position \ObjPos{} and a radius of deformation $r$. In our experiments, we set $r = 0.3$, which was determined empirically. We note that $r$ can be set adaptively based on environmental constraints and human preferences but this will be part of our future work. For waypoints within the radius of deformation $r$ from the object position \ObjPos{}, a force is applied on the waypoints in the direction of the distance vector between the waypoint and the object. The force is $0$ for other waypoints. 

For scene-independent features, a force is exerted on all waypoints of the trajectory, where the direction of the force is dependent on the feature. 

The trajectory is deformed based on the force calculated on each waypoint
:  $\DeformationFn = \InitialTraj + w\Force$, where the weight $w$ changes the magnitude of the deformation.
We empirically determined the value of $w$ to be a constant of $1.0$ in our experiments, but this can be modified based on the intensity of the language correction in the future.

Finally, the deformed trajectory is passed to a trajectory optimizer to ensure the robot's kinematic constraints are satisfied. 

\section{Experiments}
\label{sec:experiments}
We conducted simulation and real-world experiments to validate our proposed approach. We hypothesize that:

\textbf{H1} Our approach will be able to generalize to natural language phrases and environments with different objects; 

\textbf{H2} Our approach will deform trajectories at least as accurately as end-to-end methods for trajectory deformation;

\textbf{H3} Our approach will obtain higher rankings from users compared to end-to-end methods;

\textbf{H4} Our approach will be more interpretable than end-to-end methods. 

We compared our approach against LaTTe \parencite{bucker2022latte}, an end-to-end approach for language corrections that can deform trajectories in 3D space. Experiments were conducted to evaluate the generalization ability of the proposed approach. 
We also conducted user studies to evaluate the end-users' satisfaction with the deformed trajectories and to obtain more diverse language utterances.
The Nanyang Technological University Institutional Review Board approved the user studies. In the next section, we briefly describe our baseline method, LaTTe.


\subsection{Baseline}
\label{sec:baseline}

LaTTe first created a dataset consisting of tuples of initial trajectory, language correction, object images, names and poses in the scene, and deformed trajectory. This dataset is then used to train a neural network which can map the initial trajectory, language correction and objects in the scene to a deformed trajectory. For generalization to various objects in the scene and different natural language phrases, pre-trained BERT and CLIP are used to generate embeddings for the language correction and the object images, which are provided as input to the neural network. We used the model provided by the authors \footnote{https://github.com/arthurfenderbucker/LaTTe-Language-Trajectory-TransformEr}, trained on 70k samples for our experiments. Before generating the deformed trajectory using the trained model, a locality factor hyper-parameter must be set, determining the range of desired change over the trajectory. In our user studies, we set the locality factor hyper-parameter to be $0.3$, the mean value in their dataset.

\subsection{Generalization Experiments}
To evaluate the ability of our approach to generalize to different object configurations, trajectories and corrections, we used the dataset that was originally employed to train the LaTTe. The LaTTe dataset includes 100k samples, with object names sampled from the Imagenet dataset. The full dataset contains three types of trajectory modifications -- \textit{cartesian} changes, \textit{object distance} changes and \textit{speed} changes. We removed the samples related to speed changes in our evaluation as we did not include speed change features in our templates, resulting in 65261 samples. We set the locality hyper-parameter for LaTTe by referencing the value provided in each sample in the dataset.

\subsubsection{Evaluation Metrics}
\label{accuracy-measures}
To evaluate the accuracy of the deformed trajectories, LaTTe compared the similarity between the deformed trajectory using their method and the ground-truth output trajectory in the dataset using metrics like dynamic time warping (DTW) distance. 
However, this may not accurately capture the correctness of the trajectory modification. For example, even if the trajectory change occurs in a direction contrary to the intended one, the DTW distance may still register as small. Additionally, relying on a single ground-truth trajectory for comparison is problematic, as multiple valid trajectories can achieve the correct deformation. 

\begin{figure}[htpb]
    \subfloat[]{
        \label{fig: eval-cart}
        \includegraphics[trim={0.5cm 0cm 0cm 0cm},clip,width=0.23\textwidth]{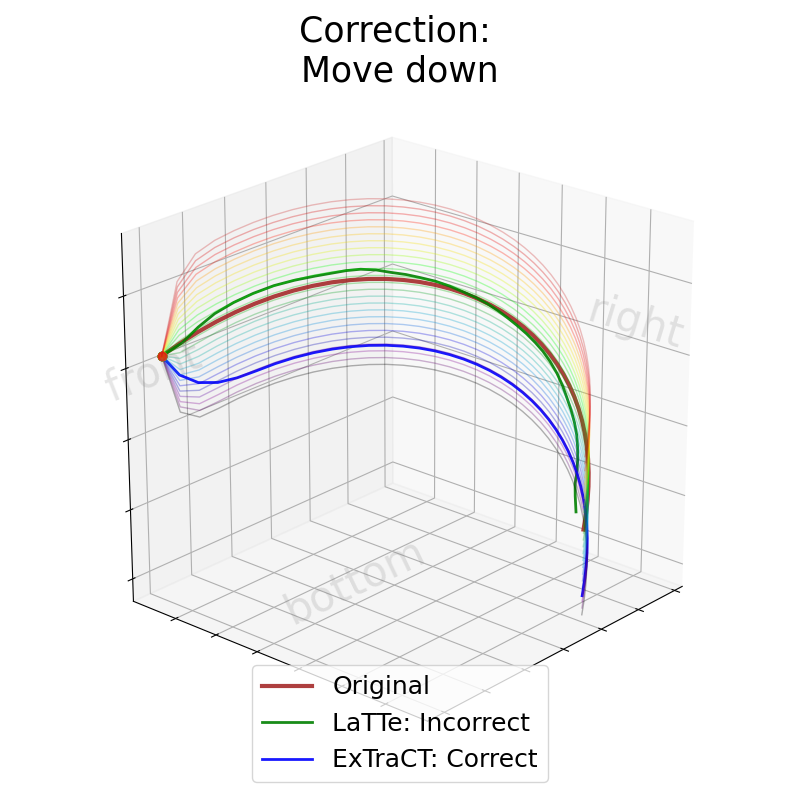}
    }
    \subfloat[]{
        \label{fig: eval-dist}
        \includegraphics[trim={0.5cm 0cm 0cm 0cm},clip,width=0.23\textwidth]{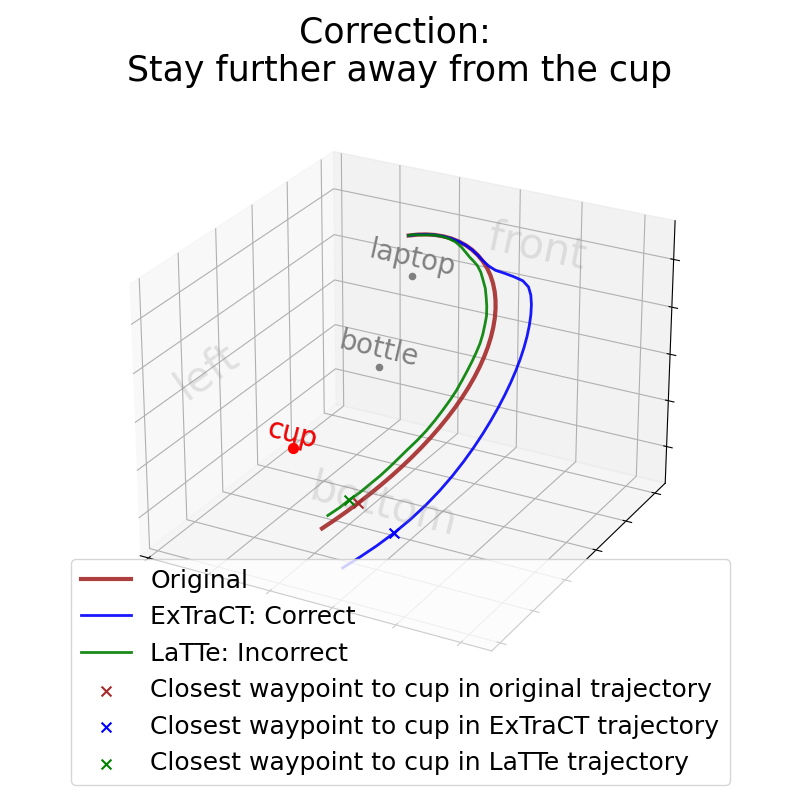}
    }
    \caption{Accuracy evaluation. The green trajectory shows an incorrect deformation, while the blue trajectory shows a correct deformation. (a) \textit{Cartesian} changes -- we sampled trajectory deformations with varying weights, which affect the intensity of deformation. The sampled trajectories below the original trajectory have negative weights, while those above the original trajectory have positive weights. (b) \textit{Object distance} changes -- we obtained the waypoints in the original and deformed trajectories and compared the change in distance relative to the target object.}
    \label{fig: eval-metrics}
\end{figure}

Thus, we developed a way to assess the performance of trajectory deformations based on language. Our approach evaluates the accuracy of deformed trajectories by considering the type of correction applied, as shown in Fig. \ref{fig: eval-metrics}. 

For corrections involving \textit{cartesian} changes, we first sampled a range of trajectory deformations of different intensities, i.e. by varying the value of $w$ from -2.5 to 2.5. 
Dynamic time warping (DTW) distance was then used to find the weight that best represents the deformed trajectory. A deformation is deemed correct if the weight aligns with the intended change direction. For example, a positive weight should correspond to a positive change (i.e. \texttt{increase}), and a negative weight should correspond to a negative change (i.e. \texttt{decrease}). 

For corrections involving \textit{object distance} changes, we measure the accuracy by comparing the distance between the object and the waypoint closest to the object in both the original and deformed trajectories. The key metric is whether the modified trajectory brings the waypoint closer to or further from the target object, which can be quantified without sampling multiple trajectories. A deformation is deemed correct if, for a positive change, the waypoint in the deformed trajectory is further from the object, and for a negative change, it is closer to the object.

To facilitate comparison with the findings reported in LaTTe, we also evaluated the performance in terms of the similarity between the deformed trajectory of each approach and the ground-truth output trajectory in the dataset using the dynamic time-warping (DTW) distance. 

\subsubsection{Results}

\begin{table*}
    \centering
    \caption{Generalization Experiment Results}
    \begin{tabular}{|c|cc|cc|cc|}
        \hline
         \multirow{2}{*}{Approach}&  \multicolumn{2}{c|}{Overall}&  \multicolumn{2}{c|}{Object Distance Changes}&  \multicolumn{2}{c|}{Cartesian Changes}\\
         &  Accuracy&  DTW&  Accuracy&  DTW&  Accuracy& DTW\\
         \hline
         ExTraCT&  \textbf{89.23\%}&  \textbf{2.6386}&  \textbf{86.10\%}&  \textbf{3.0600}&  92.27\%& \textbf{2.2100}\\
         LaTTe&  73.37\%&  3.4099&  53.48\%&  3.7499&  \textbf{92.64\%}& 3.0804\\
         \hline
    \end{tabular}
    \label{tab:gen-expt-results}
\end{table*}

\begin{figure*}[ht]
    \subfloat[]{
        \label{fig: latte-0-original}
        \includegraphics[trim={1cm 2cm 1cm 1cm},clip,width=0.23\textwidth]{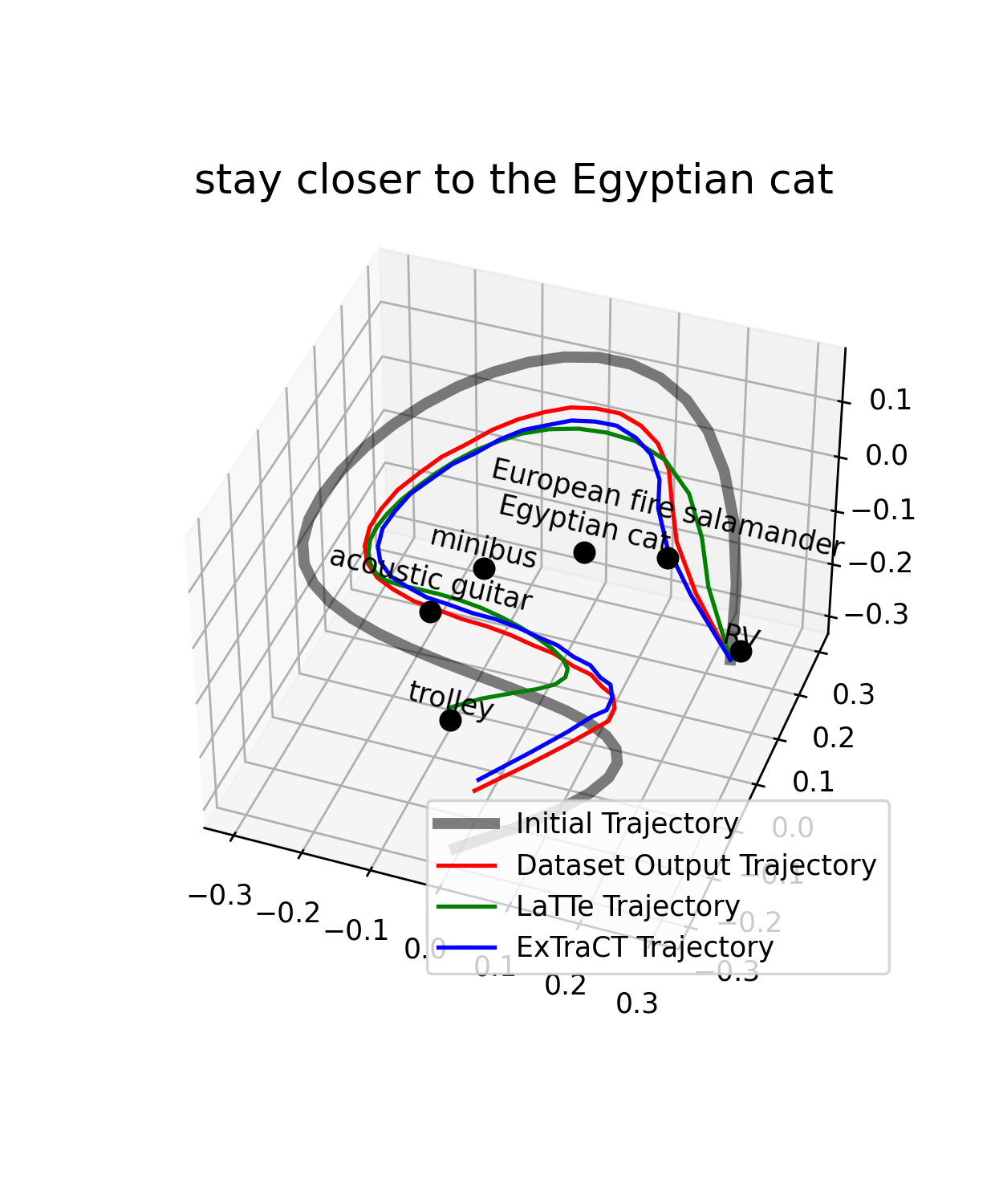}
    }
    \subfloat[]{
        \label{fig: latte-0-obj-pos}
        \includegraphics[trim={1cm 2cm 1cm 1cm},clip,width=0.23\textwidth]{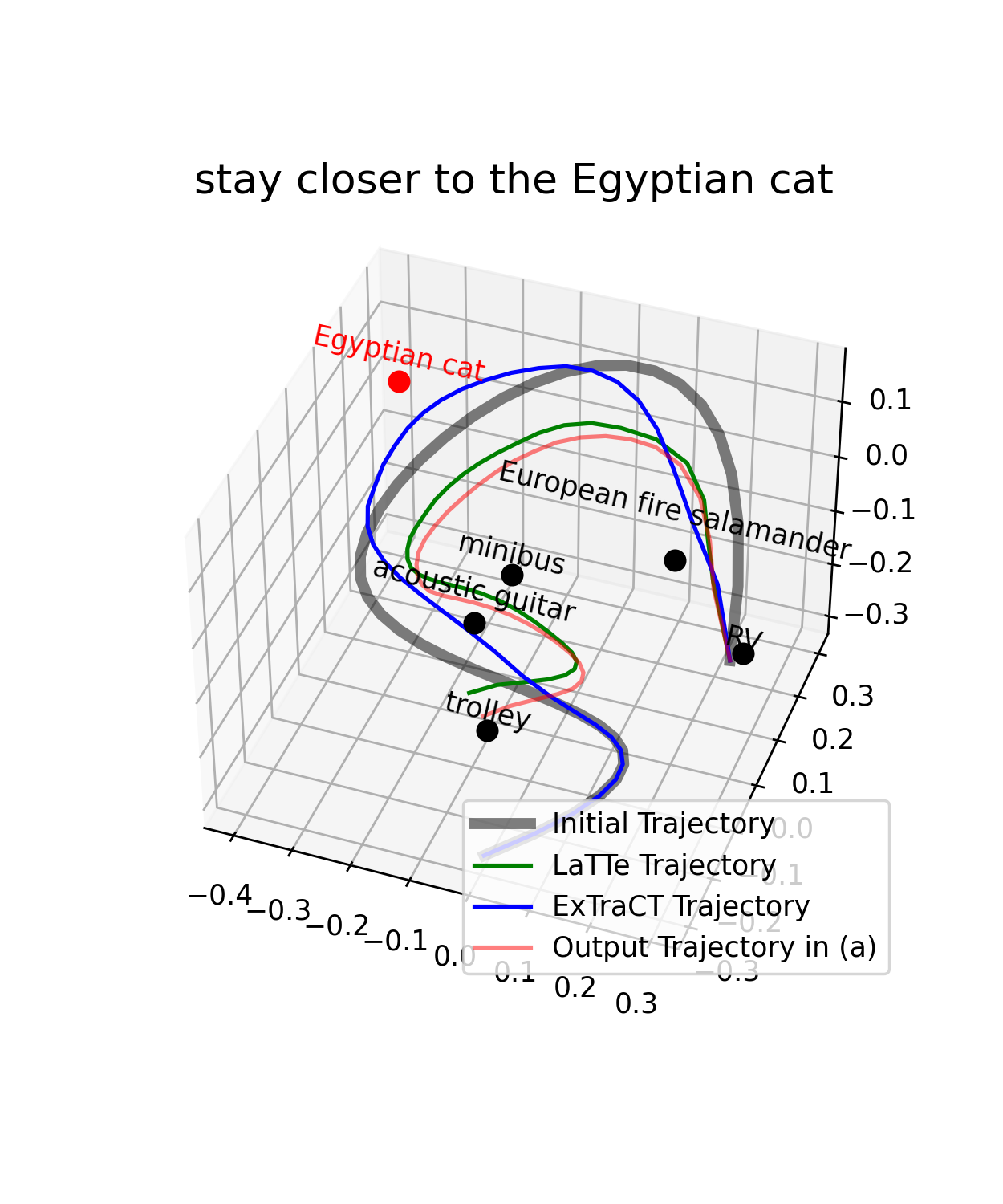}
    }
    \subfloat[]{
        \label{fig: latte-0-opp-correction}
        \includegraphics[trim={1cm 2cm 1cm 1cm},clip,width=0.23\textwidth]{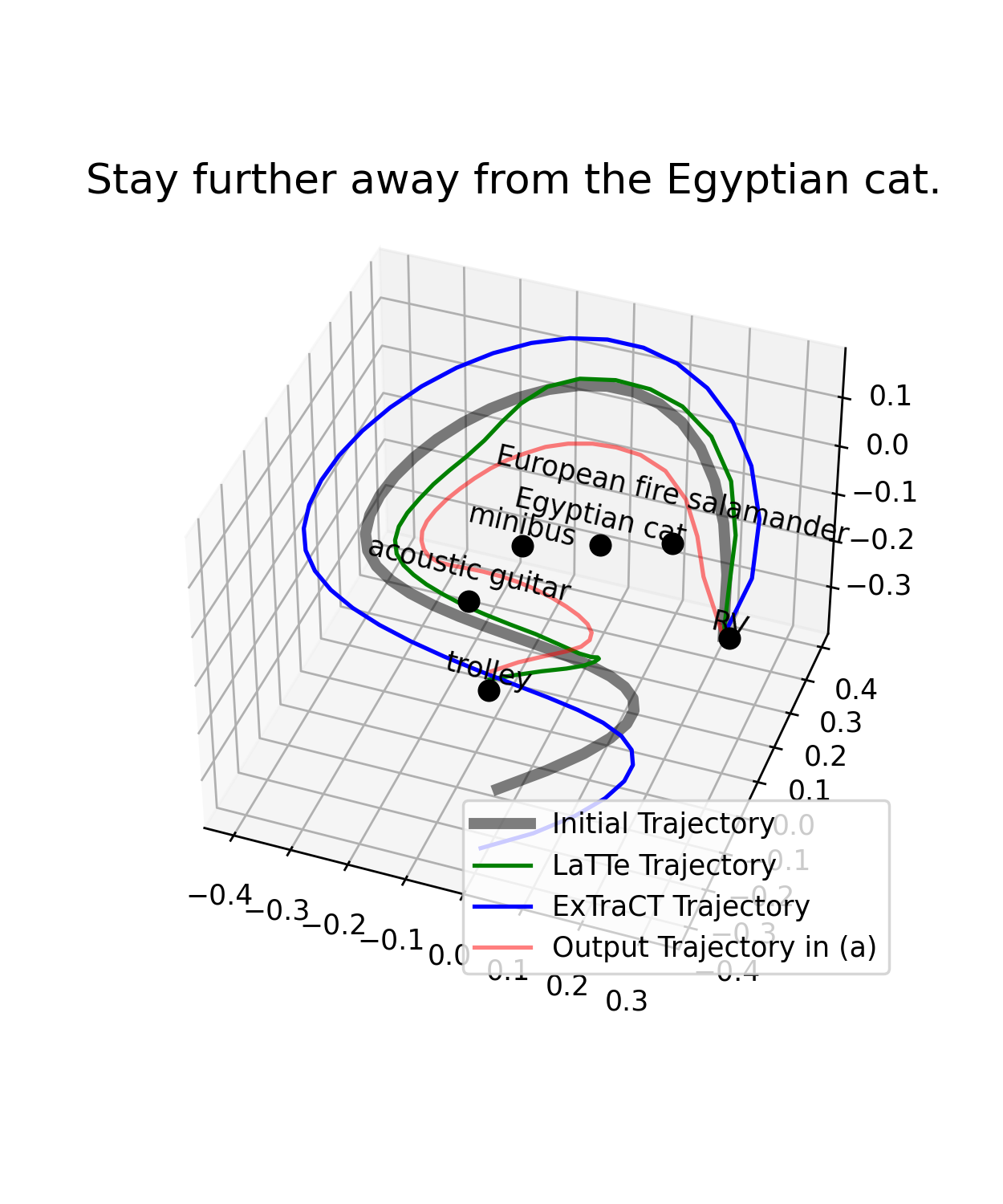}
    }
    \subfloat[]{
        \label{fig: latte-0-initial-traj}
        \includegraphics[trim={1cm 2cm 1cm 1cm},clip,width=0.23\textwidth]{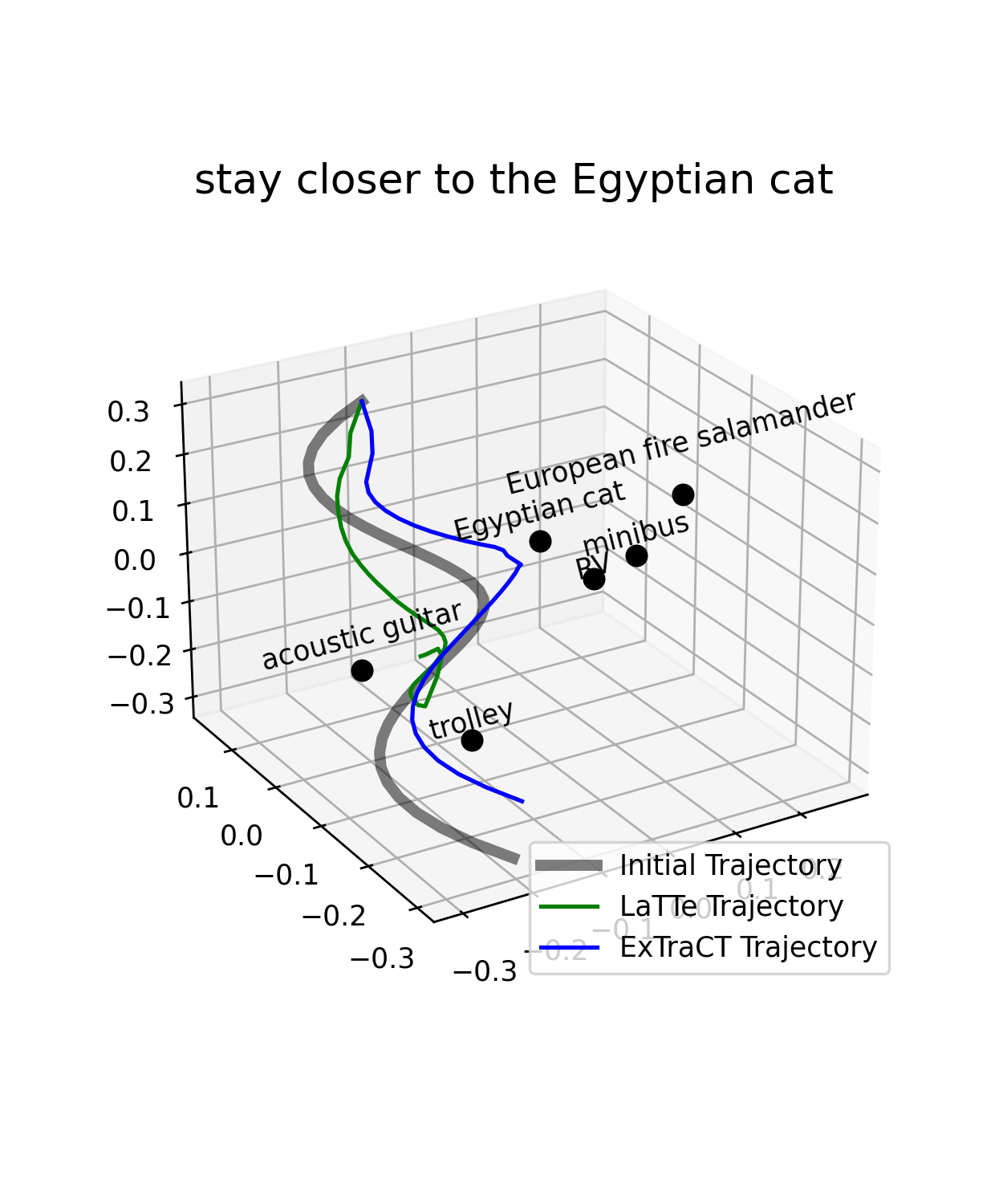}
    }
    \caption{Changes in the deformed trajectory using (a) a sample in LaTTe's dataset (b) a change in the target object pose (c) a change in the language correction that conveys an opposite meaning (d) a change in the initial trajectory. The deformed trajectory by LaTTe is inaccurate for (b), (c) and (d), while ExTraCT produces a correct trajectory deformation for all cases.}
    \label{fig: latte-analysis-0}
\end{figure*}

Table \ref{tab:gen-expt-results} shows the evaluation results. Our approach outperformed the baseline, especially for \textit{object distance} changes, showing support for \textbf{H1} and \textbf{H2}. To understand whether there is a difference in performance depending on the type of change, we analysed the results based on the type of change. Even though both approaches perform similarly for \textit{cartesian} changes, we see that the performance of LaTTe degrades for \textit{object distance} changes, with an accuracy of only \(54.5\%\), which is approximately equivalent to random chance. 

\subsubsection{Analysis of Failure Cases in LaTTe}
To better understand the degraded performance of LaTTe for \textit{distance} changes, we performed qualitative analysis on LaTTe by picking a random sample in LaTTe's dataset (Fig. \ref{fig: latte-0-original}) and modifying the target object pose (highlighted in red, Fig. \ref{fig: latte-0-obj-pos}), initial trajectory (Fig. \ref{fig: latte-0-initial-traj}) and language correction (Fig. \ref{fig: latte-0-opp-correction}). The trajectories deformed by LaTTe shown in Fig. \ref{fig: latte-analysis-0} demonstrate that these changes did not result in an expected change in the deformed trajectory. Fig. \ref{fig: latte-0-obj-pos} and \ref{fig: latte-0-opp-correction} show a similar (but incorrect) deformed trajectory to the original sample, while Fig. \ref{fig: latte-0-initial-traj} shows the deformed trajectory in the opposite direction to the correction. On the other hand, ExTraCT deforms the trajectories correctly in all these cases.

To understand why a change in the language correction (Fig. \ref{fig: latte-0-opp-correction}) did not result in a correct change in the deformed trajectory, we analysed the embeddings of the language correction. We used Sentence Transformers \parencite{reimers_2019_sentence_bert} to find the top eight sentences with the closest semantic similarity in LaTTe's dataset. Sentences that convey opposite meanings but have high lexical similarity, such as ``stay closer to the Egyptian cat'' had a high cosine similarity score to ``stay further away from the Egyptian cat'', as highlighted in Table \ref{table:sen-sim-stay-closer-egyptian-cat}. This makes learning trajectory deformations from textual embeddings difficult, as the BERT embeddings used may not always capture the semantic meaning of the corrections. 

Unfortunately, for the cases where we modified the target object pose (Fig. \ref{fig: latte-0-obj-pos}) and the input trajectory (Fig. \ref{fig: latte-0-initial-traj}), it was difficult to understand why failures occurred. Failures could arise from various factors, such as errors in embedding geometrical information like trajectories and object configurations and insufficient training data. The lack of transparency makes identifying and rectifying specific issues difficult, motivating our separation of the problem into two distinct phases -- language understanding and trajectory deformation.

\subsubsection{Analysis of Failure Cases in ExTraCT}
All the failure cases in ExTraCT can be attributed to incorrect feature mapping. For example, ``Go to the upper part" was incorrectly mapped to \texttt{z\_cartesian\_decrease} as the most similar phrase was ``Go to the bottom". ``Drive a lot closer to the meat market" was incorrectly mapped to \texttt{meat market\_distance\_increase} as the closest phrase was ``Stay a lot further away from meat market".
These errors occurred as the embeddings from large language models (LLMs) may not always capture the nuanced semantic meanings of language phrases (Table \ref{table:sen-sim-stay-closer-egyptian-cat}). 

To improve the language understanding capabilities of our system, we can either fine-tune the existing language model for our application or employ a larger LLM with better semantic understanding. Another way to improve performance is to include previously mismatched phrases in our textual description templates. In Section \ref{sec: effect-phrases}, we show how this can improve performance. 
Note that for this evaluation, we deliberately did not include all the phrases in LaTTe's dataset in our template descriptions, as that would naturally lead to an exact match in the sentence, which may not realistically reflect the model's true language understanding capabilities.



\begin{table}[htbp]
\caption{Top 8 Sentences with Highest Similarity Scores for "Stay further away from the Egyptian cat" in LaTTe's Dataset}
\begin{center}
\begin{tabular}{ | l | l | } 
    \hline
    Sentence & Similarity Score\\ 
    \hline
    Stay further away from the Egyptian cat & 1.00 \\
    Stay a lot further away from the Egyptian cat & 0.98 \\
    \textbf{Stay closer to the Egyptian cat} & \textbf{0.91} \\
    Walk a lot further away from the Egyptian cat & 0.90 \\
    Walk a bit further away from the Egyptian cat & 0.89 \\
    \textbf{Stay a lot closer to the Egyptian cat} & \textbf{0.88} \\
    \textbf{Stay very closer to the Egyptian cat} & \textbf{0.88} \\
    \textbf{Stay a bit closer to the Egyptian cat} & \textbf{0.87} \\
    \hline
\end{tabular}
\begin{tablenotes}
    Bolded phrases indicate sentences with opposite meanings.
\end{tablenotes}
\end{center}
\label{table:sen-sim-stay-closer-egyptian-cat}
\end{table}

\subsection{User Studies in Simulation}

\begin{figure}[t]
    \centering
    \includegraphics[width=0.8\linewidth]{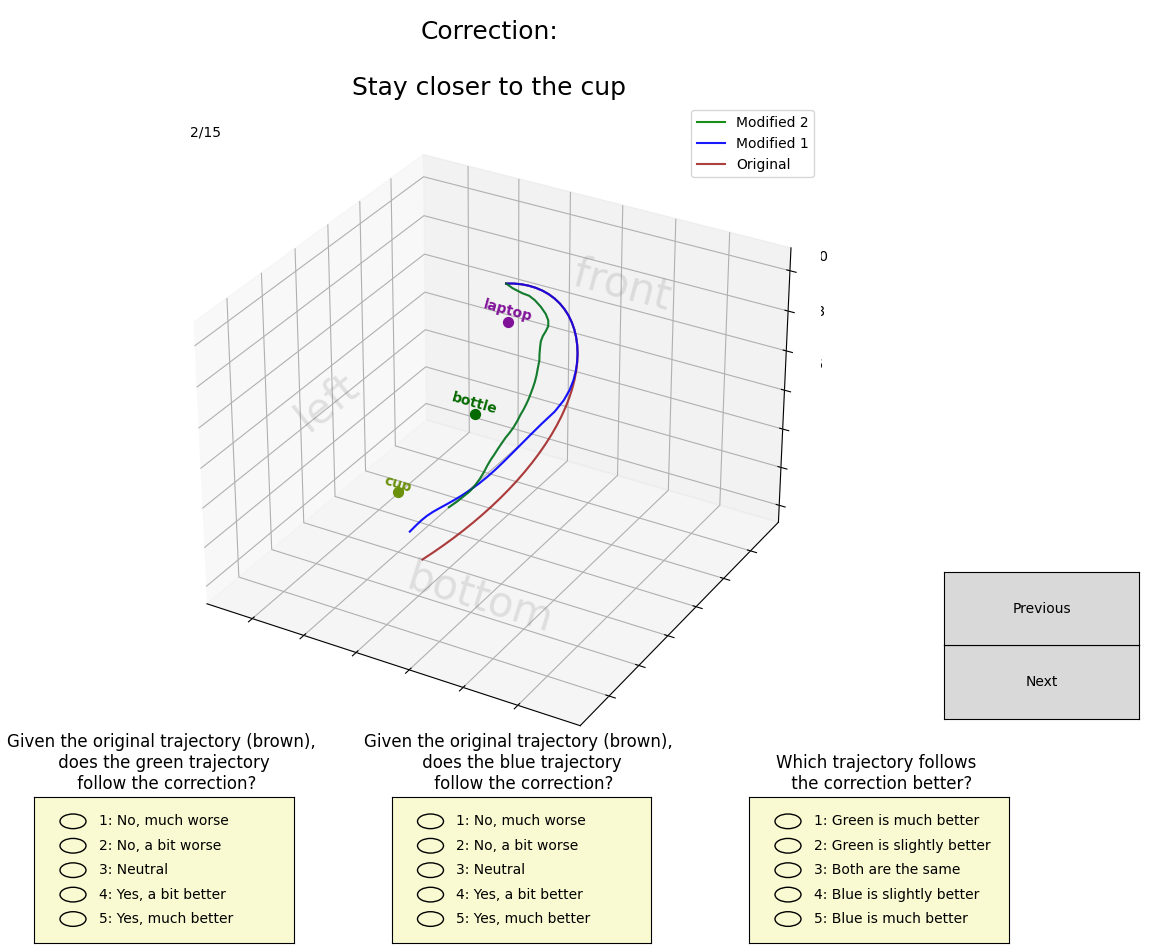}
    \caption{Interface for the simulated study showing scene 1. The modified trajectories are displayed on the interface simultaneously for better comparison. The green modified trajectory is by LaTTe, while the blue modified trajectory is by our approach.}
    \label{sim-scene1}
\end{figure}

We recruited 15 subjects (8 male, 7 female) for the simulation study. Out of these participants, 6 individuals did not have a background in robotics. The study interface (Fig. \ref{sim-scene1}) (modified from LaTTe), displayed the initial trajectory, the modified trajectory using LaTTe and the modified trajectory using our method (ExTraCT) at the same time. The trajectories were displayed to the users in 3D and subjects could interact with the plots to change the view. 
The participants were not informed which method was used to deform the trajectories, and the modified trajectories were labelled only with "1" and "2".
The labels were kept consistent throughout the experiment so that we could obtain the subjects' overall rankings and preferences for each method across different scenes. 



\begin{figure}[!t]
\centering
    \subfloat[\label{sim-scene2}]{
        \includegraphics[width=0.5\linewidth, trim={2cm, 1cm, 0.5cm, 0.8cm}, clip]{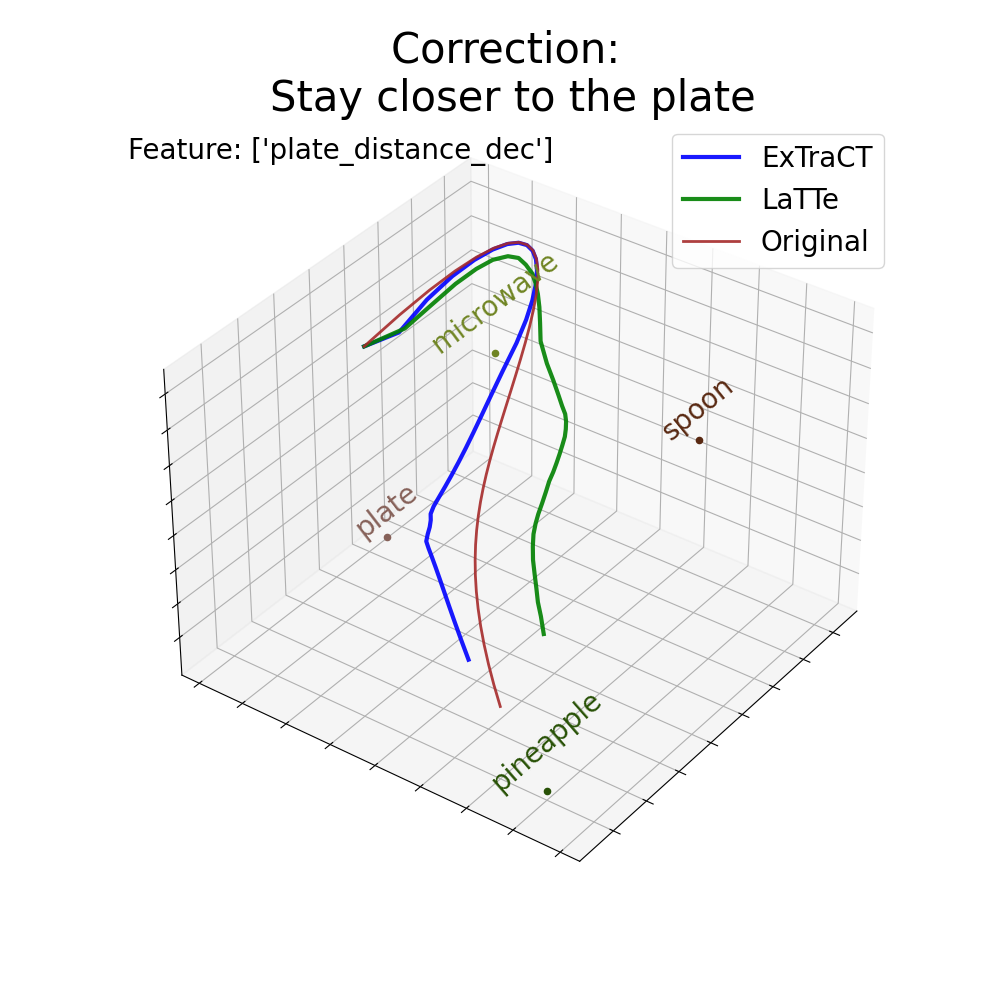}
    }
    \subfloat[\label{sim-scene3}]{
        \includegraphics[width=0.5\linewidth, trim={2cm, 1cm, 0.5cm, 0.8cm}, clip]{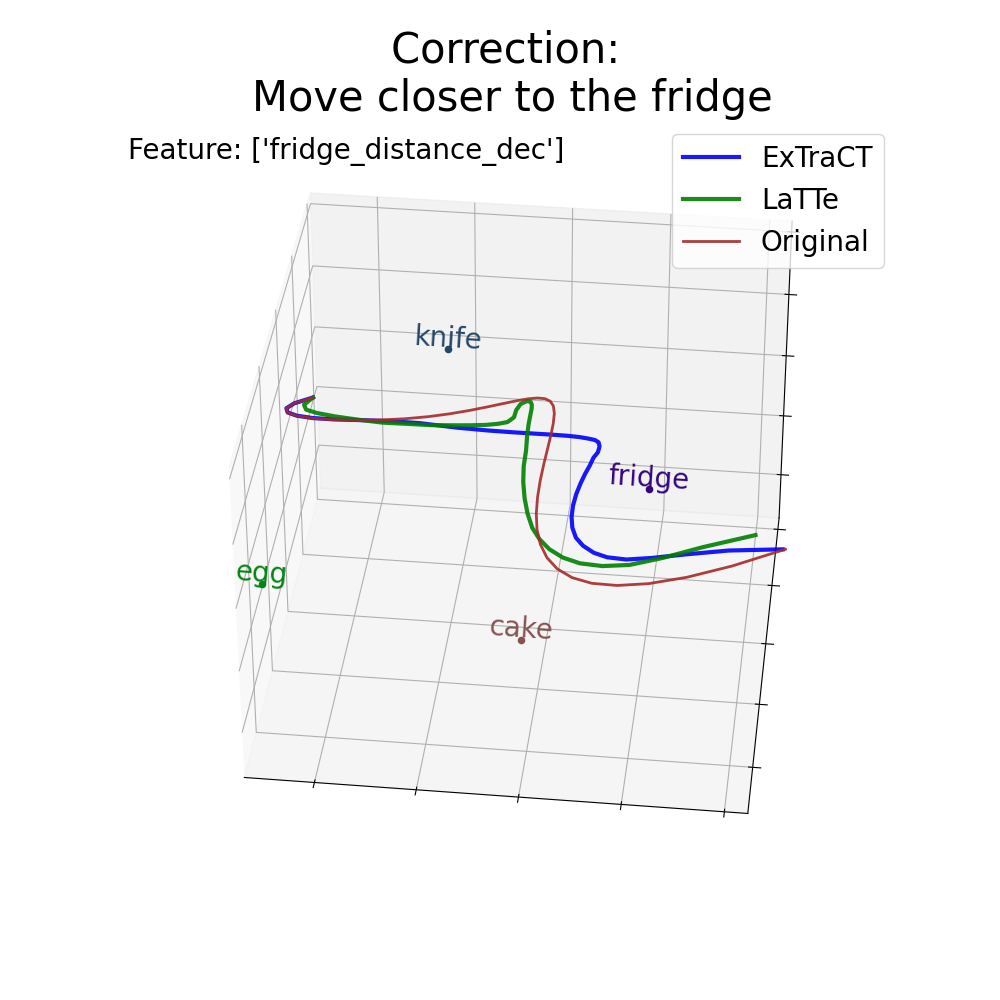}
    }
\caption{The deformed trajectories and features matched for (a) scene 2 and (b) scene 3. Note that the modified trajectory by LaTTe opposes the corrections provided in these examples.}
\label{fig:sim_study_ss}
\end{figure}

3 scenes were presented to each subject, as shown in Figures \ref{sim-scene1} and \ref{fig:sim_study_ss}. Different household objects were placed at randomized locations for each scene. The objects in scenes 1 and 2 were selected from LaTTe's dataset, while those in scene 3 were randomly selected and out of LaTTe's dataset.
Since there were no images of the objects, no CLIP image embeddings were used for LaTTe. Instead, CLIP textual embeddings of the object names were used for LaTTe to identify the correct target object.

There were 5 language corrections for each scene, where 3 of the corrections were given by us (for consistency across subjects and to familiarize subjects with the types of corrections), and 2 of the corrections were given by the subjects. Subjects were free to provide any language corrections, after which both approaches would modify the trajectory. The corrections we provided were in LaTTe's dataset for scenes 1 and 2, and were not present in LaTTe's dataset for scene 3. After each correction, subjects had to rank their agreement on how well each method modified the trajectory and which trajectory they preferred, if any. Specifically, for each method, they had to choose whether the deformed trajectory was 1) Completely wrong, 2) Somewhat wrong, 3) Neutral, 4) Somewhat correct, or 5) Completely correct. After each trajectory deformation and at the end of the experiments, they had to compare both methods. For this, they had to choose whether 1) Method 1 was much better, 2) Method 1 was a bit better, 3) Both the methods were the same, 4) Method 2 was a bit better, or 5) Method 2 was much better. At the end of the study, subjects were asked to elaborate on which method they preferred and why. We also measured the performance based on the accuracy of the trajectory deformations, as outlined in Section \ref{accuracy-measures}.

\subsubsection{Results}

\begin{table*}
    \centering
    \caption{Simulation User Study Results}
    \begin{tabular}{|c|cc|cc|cc|}
        \hline
         \multirow{2}{*}{Approach}&  \multicolumn{2}{c|}{Overall}&  \multicolumn{2}{c|}{Object Distance Changes}&  \multicolumn{2}{c|}{Cartesian Changes}\\
         &  Accuracy&  Mean User Rank&  Accuracy&  Mean User Rank&  Accuracy& Mean User Rank\\
         \hline
         ExTraCT &  \textbf{88.00\%} & \textbf{4.471} $\pm$ 0.062 & \textbf{96.53\%} &  \textbf{4.521} $\pm$ 0.0668 & \textbf{100.00\%} & \textbf{4.776} $\pm$ 0.060 \\
         LaTTe &  $55.56\%$ & $2.840 \pm 0.0790$ & $47.22\%$ & $2.722 \pm 0.0988$ & $94.83\%$ & $3.207 \pm 0.136$ \\
         \hline
    \end{tabular}
    \label{tab:sim-expt-results}
\end{table*}

Table \ref{tab:sim-expt-results} shows the results of our user study, showing better accuracy and preference for our method, supporting \textbf{H2} and \textbf{H3}. A Wilcoxon signed-rank test showed a significant difference in the rankings between our method and LaTTe ($p<0.0001$) on how well the trajectory followed the correction. When comparing the two methods after each correction, subjects rated that our method was better $85.5\%$ of the cases, LaTTe was better $6.2\%$ of the cases, and there was no difference for $8.0\%$. When asked at the end of the experiments, all the subjects preferred our method, with $33.3\%$ of subjects rating that our method was ``slightly better'', and $66.7\%$ of subjects rating that our method was ``much better'' than LaTTe. 

The higher preference for our method could be largely attributed to the accuracy of our approach, with many subjects stating that ExTraCT ``follows [the] correction more''. LaTTe did not deform trajectories according to the corrections for many cases, such as in Fig. \ref{sim-scene2} and \ref{sim-scene3}, where the deformed trajectories were opposite to the corrections.

\subsubsection{Failure Cases}
There were $12.00\% $ of failure cases with ExTraCT. Due to the explainability of our approach, we can analyse the feature matching and confidence score for each failure case (\textbf{H4}). We labelled the correct feature(s) for corrections within our feature space. For example, ``Move below the fridge", which contains a directional component for an object, is out of our feature space. 
These failures can be attributed to a lack of a correct deformation function.
There were 12 such corrections ($5.33\%$), which we removed from subsequent analysis. 

The remaining 15 failures could be attributed to two reasons. Of these, 10 failures were due to multiple trajectory modification features within a single correction, such as ``Move away from bottle and then closer to cup". Since we assumed that each correction only contains one feature, such failures were expected. 5 failures were due to incorrect feature mapping, i.e. ``Stay further away from the pineapple'' was incorrectly mapped to \texttt{pineapple\_distance\_decrease}, as the most similar phrase was ``Stay close to pineapple''; and ``Move to the cake'' was mapped to \texttt{cake\_distance\_increase} since the most similar phrase was ``Move further away from cake''. 
These failures highlight the limitations of using Sentence Transformers for capturing semantic nuances in language corrections. Fine-tuning the model with in-domain data could improve the performance. 

\subsubsection{Effect of textual description templates on performance}
\label{sec: effect-phrases}

We investigated the impact of the size of textual descriptions (number of phrases in each textual description) and phrase selection on performance. We selected phrases less frequently provided in the simulation experiments from the original set of textual description templates. The resulting textual description templates contained only 2 phrases, highlighted in bold in Table \ref{table:features}. Using this smaller set of phrases, we examined the features matched for the 59 unique language corrections from the simulation study. With a smaller set of language phrases, the number of errors in terms of feature mapping remained at 6, but there were 3 instances of low confidence. Even though the number of errors was the same, the phrases with incorrect feature mapping differed.

We also demonstrate how we can easily add phrases to our textual description templates and improve the system's performance. We added the following phrases that were incorrectly mapped to the feature during the simulation study -- ``Stay further away from the \{obj\}'' to \texttt{obj\_distance\_inc}; ``Move towards \{obj\}'', ``Move to \{obj\}'' to \texttt{obj\_distance\_dec}; and ``Move to the right'' to \texttt{y\_cart\_inc}.
After adding the phrases with incorrect feature mappings to the textual description templates, there were no more inaccurate feature mappings.

From this, we can conclude that while using a pre-trained LLM for semantic similarity matching can capture some complexity in natural language, they do not generalize to all cases due to the diversity and flexibility of natural language.
The explainability of our approach provides valuable insights to enhance the system's performance.
We demonstrate that language grounding to features can be improved by expanding the phrases in the textual description templates, improving our approach's generalizability to language variations. 



\begin{figure}[!htb]
\centering
    \subfloat[\label{real-scene1}]{
        \centering
        \includegraphics[width=0.32\linewidth, trim={2cm, 2cm, 2cm, 0.8cm}, clip]{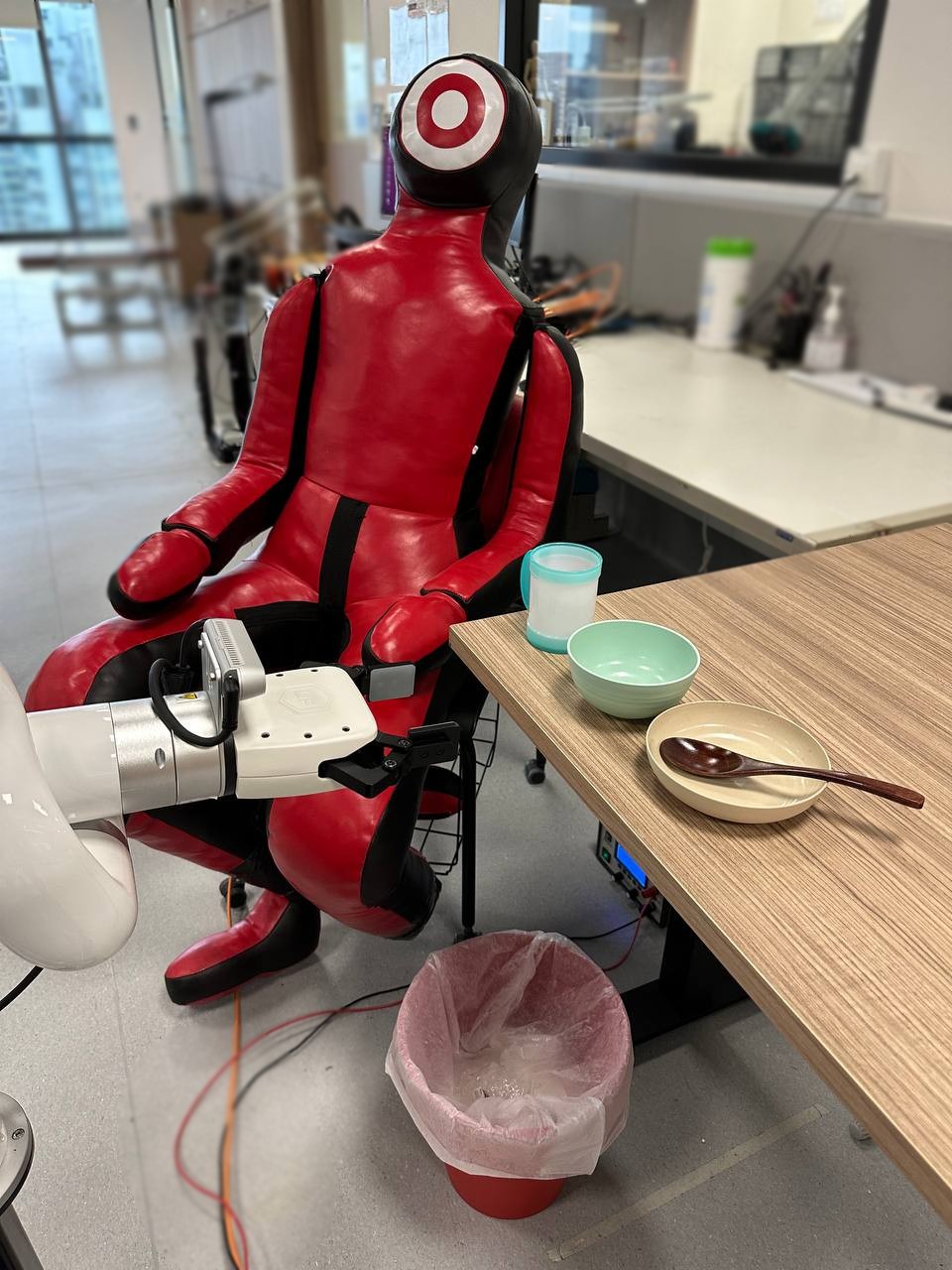}
    }
    \subfloat[\label{real-scene2}]{
        \centering
        \includegraphics[width=0.32\linewidth, trim={2cm, 2cm, 2cm, 0.8cm}, clip]{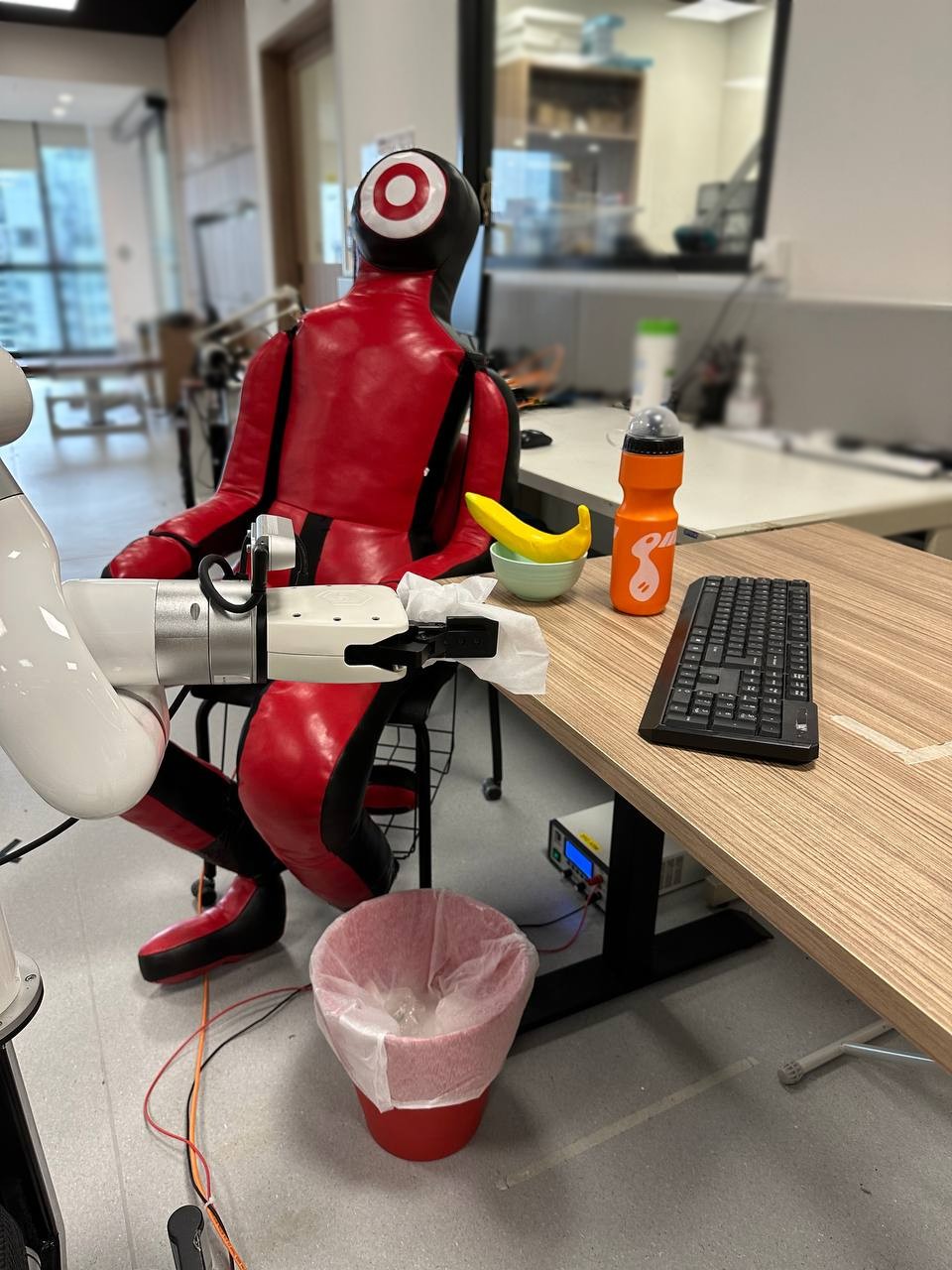}
    }
    \subfloat[\label{real-scene3}]{
        \centering
        \includegraphics[width=0.32\linewidth, trim={2cm, 2cm, 2cm, 0.8cm}, clip]{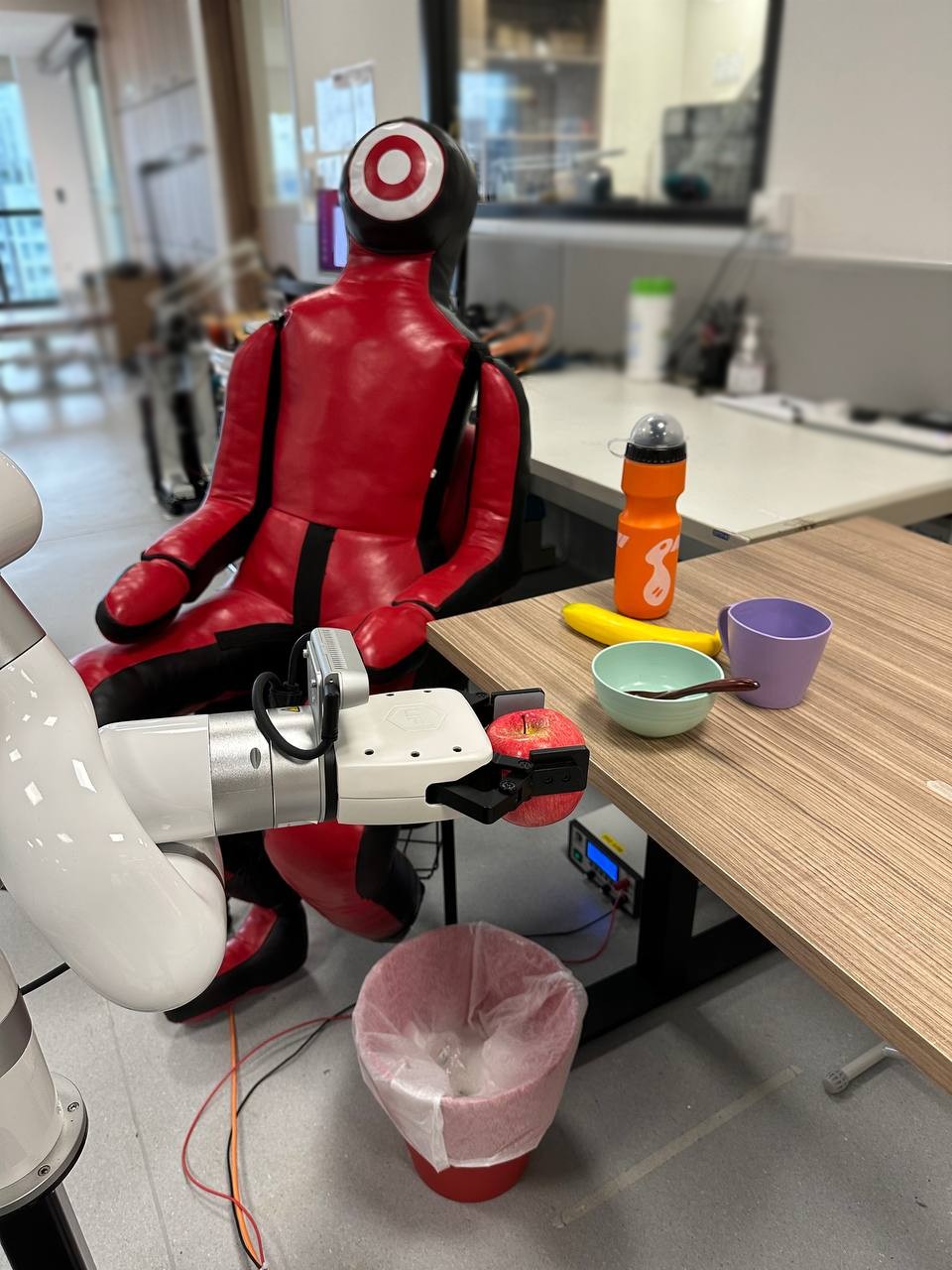}
    }
\caption{Setup of the experiments using the xArm-6 manipulator for (a) scene 1 (b) scene 2 and (c) scene 3. For safety, users were seated outside the robot's workspace and a dummy was placed at the scene.}
\label{fig:real_expt_setup}
\end{figure}

\subsection{User Studies with Real Arm}
We also conducted user studies using the xArm-6 robotic arm (UFactory, China) with the xArm two-finger gripper. An Intel Realsense Depth Camera D435 was mounted on the xArm-6 gripper to capture images of the scene to estimate the location of the objects. Additionally, for LaTTe, the bounding boxes of the objects were also estimated to obtain the CLIP embeddings of the objects. Object detection was performed using Mask R-CNN \parencite{he2017mask} in our experiments. TrajOpt \parencite{schulman2014motion} was used to optimize the trajectory based on the robot kinematic constraints.

We recruited 5 subjects (2 male, 3 female) for a within-subject study, where each subject evaluated two methods -- our approach (ExTraCT) and LaTTe. 
\subsubsection{Setup and Protocol}
There were 3 scenes with different objects on the tables. Fig. \ref{fig:real_expt_setup} shows the scenes used for real arm experiments. The object types, number of objects and object positions varied across the scenes. For the first scene, we provided the language corrections for consistency across subjects and also to familiarize the subjects with the types of corrections. The subjects had to rate the performance of the following corrections for both methods --  "Go down", "Keep a bigger distance away from bowl", "Stay closer to spoon". 

For the next 2 scenes, subjects had to use language corrections to modify a trajectory to complete tasks. Subjects were informed that they could only give 1 correction at a time, with 3 corrections required. The 2nd scene required subjects to throw trash into the bin while avoiding the food items and objects on the table. The 3rd scene required subjects to bring an apple to a dummy for handover while staying away from the bowl and staying closer to the table. In both scenes, subjects also had to modify the trajectory's final waypoint to move closer to the bin or the dummy. The language corrections were manually typed into the computer running both methods. Subjects completed both methods for each scene before moving to the next scene. 

The order of the methods was randomized across subjects to counteract the effects of novelty and practice. Subjects filled out a qualitative survey after each task and each method, rating on a scale of 1 to 5 whether the modified trajectory followed the language correction. 
At the end of the study, subjects were asked to elaborate on their preferred method and why. We also measured the accuracy of the trajectory deformations using the same method in Section \ref{accuracy-measures}.

\subsubsection{Results}
The real-arm study showed similar results to the simulation study (Table \ref{table:real-arm-expt-result}), where ExTraCT deformed trajectories more accurately than LaTTe (\textbf{H2}). A Wilcoxon signed-rank test was performed on the user rankings. 
Statistically significant differences were found, with our trajectory deformations following the corrections better ($p < 0.0001$), showing support for \textbf{H3}. Subjects generally preferred our method, with $80\%$ of subjects rating our method slightly better or much better than LaTTe. Subjects preferred our method because it ``captures the user's intention better" and has a ``greater accuracy". 



    

\begin{table}[!h]
\caption{Real Arm Experiment Results}
\begin{center}
\begin{tabular}{ | c | c | c | } 
    \hline
    Method & Accuracy & Mean User Rank \\ 
    \hline
    ExTraCT & \( \textbf{97.78}\% \)& \(\textbf{4.38} \pm 0.12\) \\
    LaTTe & \( 66.67\% \)& \(3.16 \pm 0.20\) \\
    \hline
    
\end{tabular}
\end{center}
\label{table:real-arm-expt-result}
\end{table}

\subsection{Application in Assistive Feeding}
\label{sec:feeding}
\begin{figure}[htbp]
    \centering
    \subfloat[\label{fig: feeding-setup}]{
        \centering
        \includegraphics[height=5cm, 
        ]{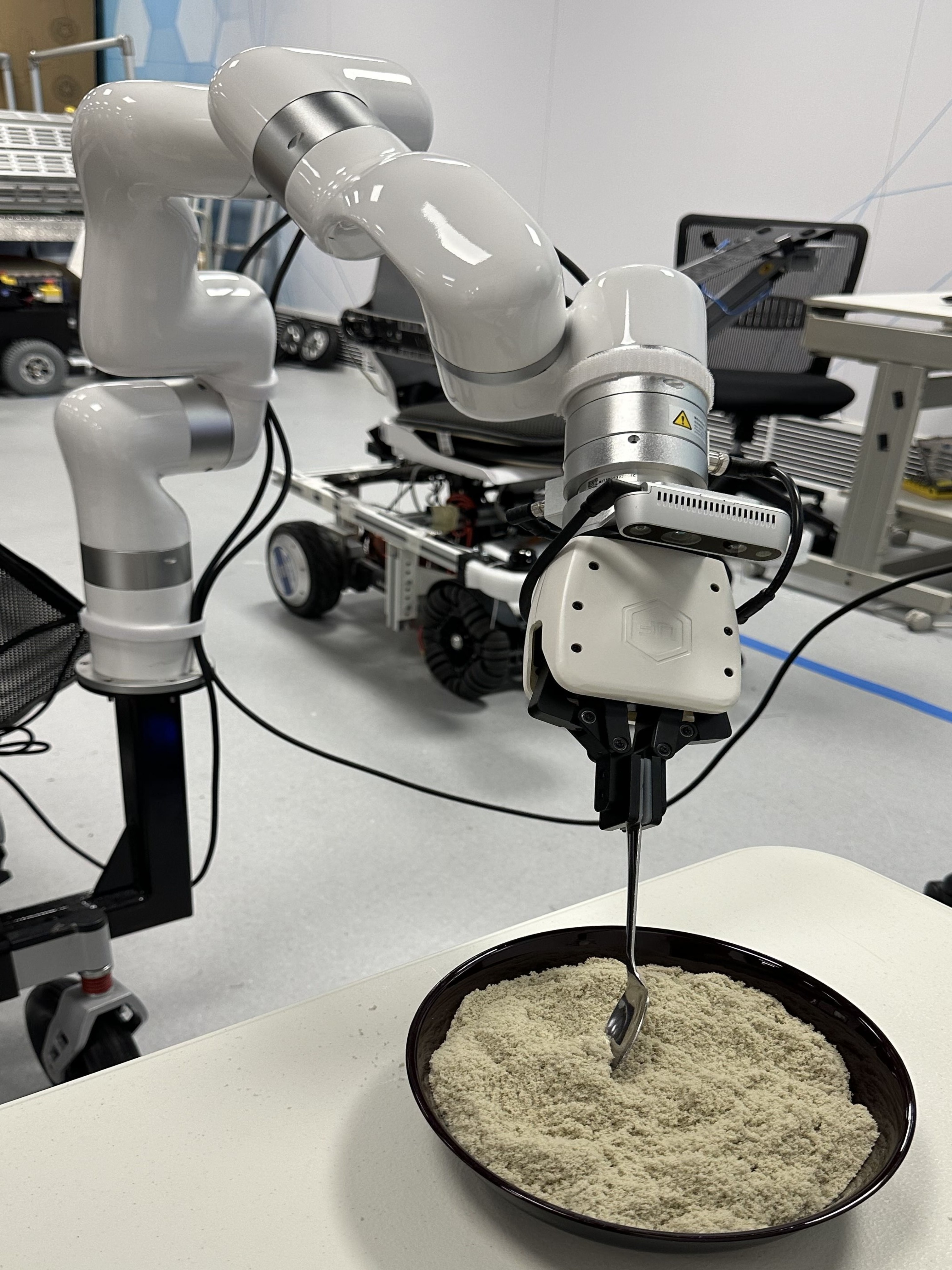}
    }
    \subfloat[\label{fig: feeding-traj}]{
        \centering
        \includegraphics[height=5cm, 
        ]{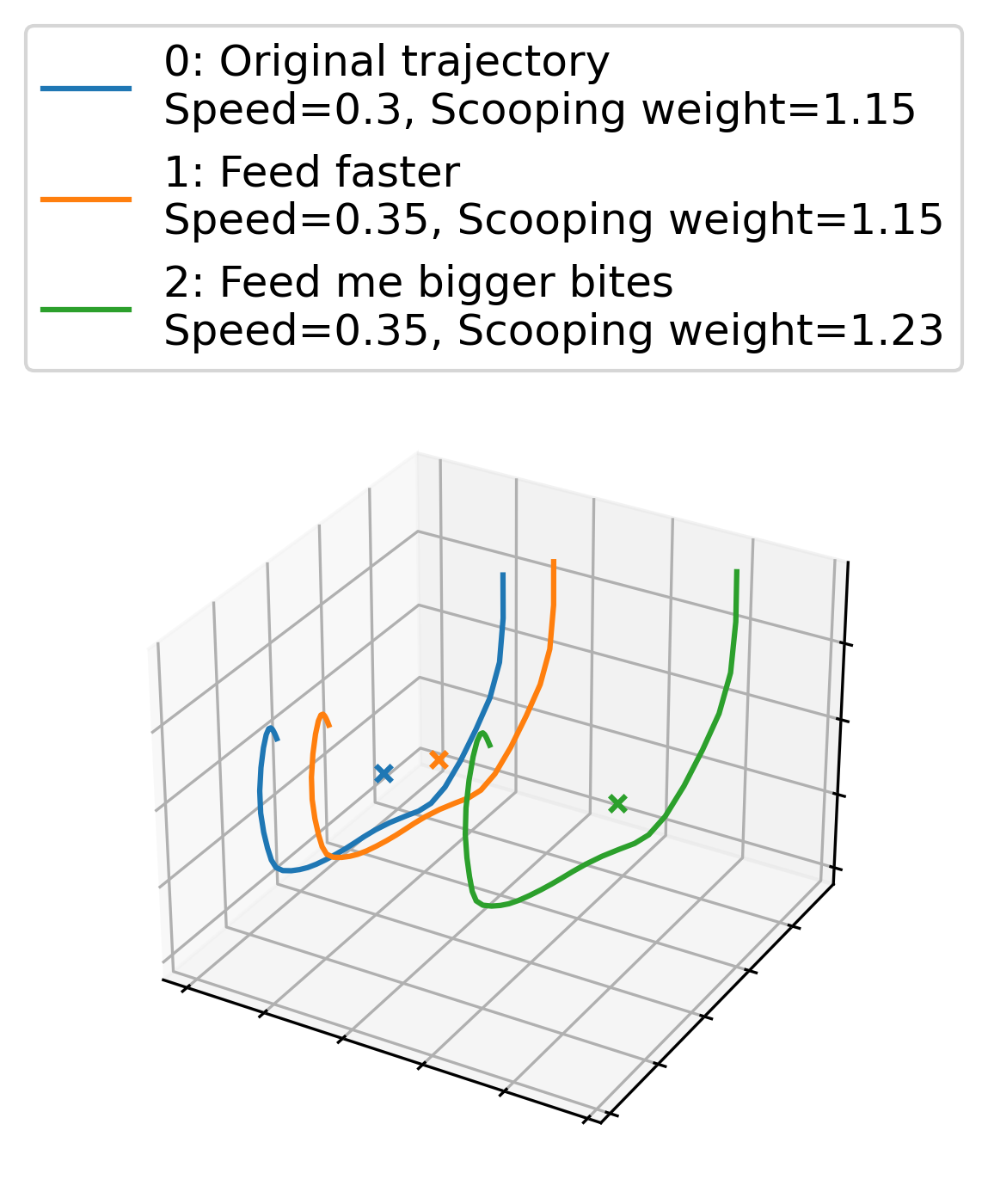}
    }
    \newline
    \subfloat[\label{fig: original-bite-size}]{
        \centering
        \includegraphics[height=4cm]{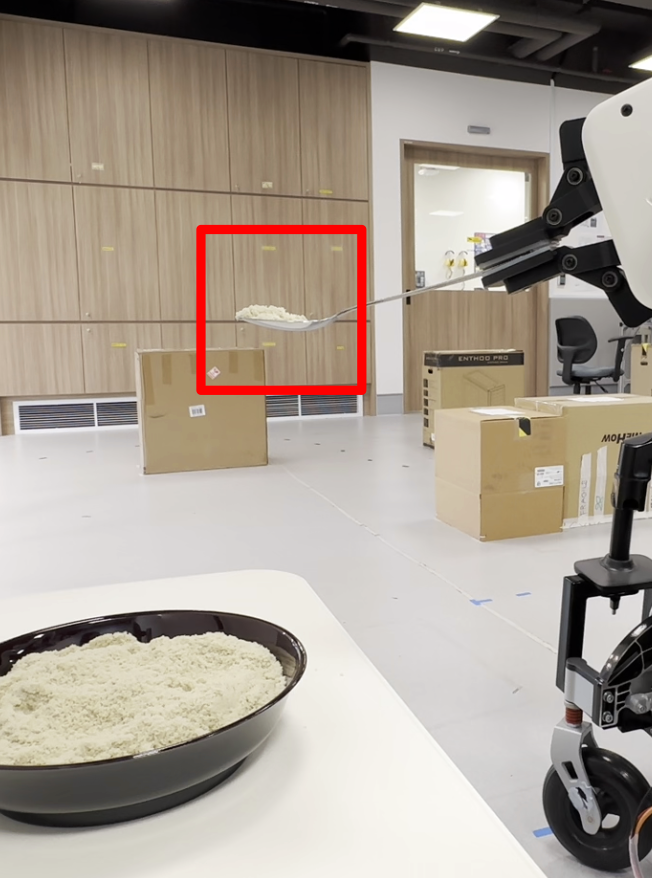}
    }
    \subfloat[\label{fig: larger-bite-size}]{
        \centering
        \includegraphics[height=4cm]{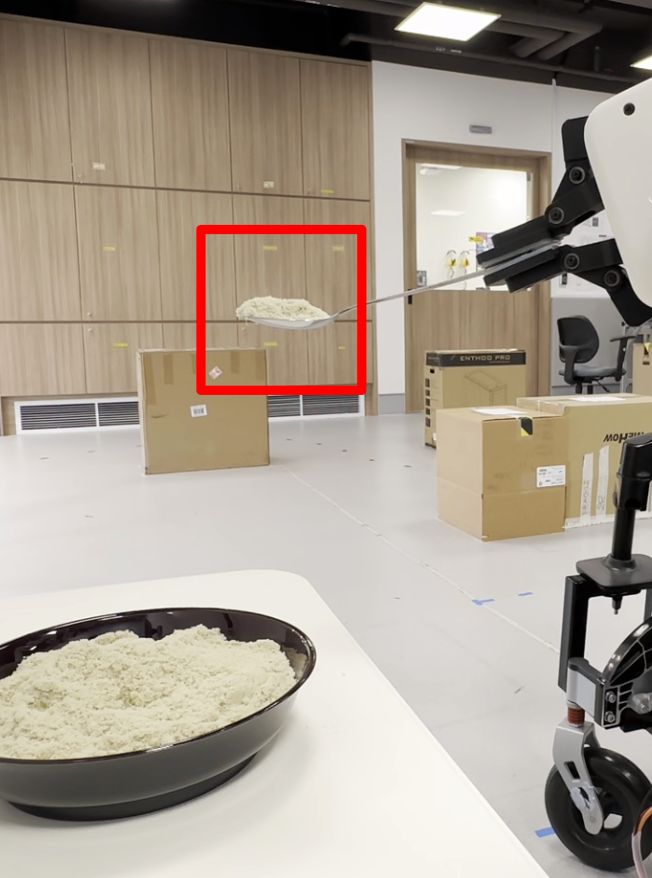}
    }
    \caption{(a) Setup for assistive feeding using a spoon. (b) The change in scooping trajectories after a language correction is provided. (c) Original bite size. (d) Bite size after the correction ``Feed me bigger bites" was provided.}
    \label{fig: feeding}
\end{figure}

We deployed our framework in an assistive feeding task to show that our framework is versatile and can be applied across diverse scenarios. 
To tailor our framework for this context, we defined two features -- \textit{bite size} and \textit{feeding speed}. The features and textual descriptions are shown in Table \ref{table:features-feeding}. We show some sample interactions on how the amount of food scooped can be modified using language in Fig. \ref{fig: feeding}. The modularity of our framework allows different trajectory modification methods to be used depending on the task's specific requirements. In this case, we used parameterized dynamic motion primitives (DMP) to modify the bite size.

Fig. \ref{fig: feeding-traj} displays the variations in scooping trajectories based on language corrections. After the correction ``Feed faster" was provided, the scooping speed increased. Similarly, following the correction ``Feed me bigger bites", the scooping trajectory was modified by scaling up the weight of the DMP, resulting in a larger bite size, as seen in Fig. \ref{fig: larger-bite-size}.

\begin{table}[!h]
\caption{Features (F) and their Textual Descriptions (TD) for Assistive Feeding}
\begin{center}
\begin{tabular}{ | m{0.5cm} | m{3.5cm} | m{3.5cm} | } 
    \hline
    F & {bitesize\_decrease} & {bitesize\_increase}\\ 
    \hline
    \multirow{5}{*}{TD} & I want a bigger bite & I want a smaller bite \\
    & Increase the spoonful size & Decrease the spoonful size \\
    & I want a larger bite next & I want a smaller bite next \\
    & I want more food & I want less food \\
    & Increase bite size & Decrease bite size\\
    \hline
    
    \multicolumn{3}{c}{} \\[-0.5em] 
    
    \hline
    F & {speed\_decrease} & {speed\_increase} \\
    \hline
    \multirow{3}{*}{TD} & Move faster & Move slower \\
    & Increase speed & Decrease speed \\
    & Too slow & Too fast \\
    \hline
    
    \multicolumn{3}{c}{} \\[-0.5em]
    
\end{tabular}
\end{center}
\label{table:features-feeding}
\end{table}

\section{Conclusion and Discussion}
\label{sec:conclusion}
In conclusion, we propose a modular trajectory correction framework, ExTraCT. ExTraCT creates more accurate trajectory modification features for natural language corrections, which is important for safety and building trust in human-robot interaction. Our proposed architecture uses pre-trained LLMs for grounding user corrections and semantically maps them to the textual description of the features. 

Our approach combines the strengths of hand-crafted features in trajectory deformations to generalize to different object configurations and initial trajectories and the language modelling capabilities of LLMs to handle language variations.
By separating the problem of language understanding and trajectory modification, we have shown improvements in interpreting and executing language corrections, even for non-templated language phrases. The transparency of our approach allowed us to understand the root causes of failures, whether in language understanding or the trajectory deformation process, enabling more targeted improvements to the system.

This work is just a step towards understanding how explicitly obtaining the features for trajectory modification can help provide a more explainable and generalizable approach. Our feature space is limited, and we cannot extract multiple features from a single correction. While our current implementation allows a chain of corrections, correcting the trajectory one feature at a time may not be efficient, especially since natural language can handle greater complexity. Future work aims to increase our feature space and handle more complex language utterances, such as different intensities of trajectory modifications, compound sentences and referring expressions.
Another direction is to look into bi-directional communication between the robot and the user, allowing the robot to ask the user for clarifications in cases of uncertainty (i.e. low confidence in feature mapping).
We hope our framework can provide a more transparent approach to learning human preferences, which can serve as a basis for transferring human preferences across different contexts.

\section*{Ethics Statement}
The studies involving humans were approved by the Nanyang Technological University Institutional Review Board. The participants provided their written consent to participate in the studies.



\section*{Funding}
This work is supported by the Rehabilitation Research Institute of Singapore and the National Research Foundation, Prime Minister's Office, Singapore, under its Campus for Research Excellence and Technological Enterprise (CREATE) programme.

\section*{Acknowledgments}
The authors acknowledge the use of OpenAI's ChatGPT in the editing process of the paper.


\printbibliography


\end{document}